\documentclass{article}

\usepackage{arxiv}
\usepackage[square,numbers]{natbib}
\usepackage{mypackages}
\usepackage{mymacros}
\usepackage{myacronyms}
\title{Integrating Machine Learning into\\Belief-Desire-Intention Agents: Current\\Advances and Open Challenges}

\date{}

\author{Andrea Agiollo \\
        \tudelft \\
        Van Mourik Broekmanweg 5\\
        Delft, ZH, The Netherlands, 2628 XE\\
        \texttt{A.Agiollo-1@tudelft.nl}
        \And
        Andrea Omicini\\
        \unibo\\
        via dell'Università 50\\
        Cesena, FC, Italy, 47522\\
        \texttt{andrea.omicini@unibo.it}
}

\renewcommand{\shorttitle}{ML-BDI Agents: Current Advances and Open Challenges}
\renewcommand{\undertitle}{This manuscript is currently under review at ACM Computing Surveys.}
\renewcommand{\headeright}{Under Review}

\begin{document}
\maketitle

\begin{abstract}
    Thanks to the remarkable human-like capabilities of \ac{ML} models in perceptual and cognitive tasks, frameworks integrating \ac{ML} within \emph{rational agent architectures} are gaining traction. 
    Yet, the landscape remains fragmented and incoherent, often focusing on embedding \ac{ML} into generic agent containers while overlooking the expressive power of rational architectures\textemdash{}such as \emph{\ac{BDI} agents}. 
    This paper presents a fine-grained systematisation of existing approaches, using the \ac{BDI} paradigm as a reference. 
    Our analysis illustrates the fast-evolving literature on rational agents enhanced by \ac{ML}, and identifies key research opportunities and open challenges for designing effective rational \ac{ML} agents.
\end{abstract}

\section{Introduction}\labelsec{intro}

Intelligent agents designed around \emph{explicit} mentalistic notions \textendash{} such as \emph{beliefs} and \emph{goals} \textendash{} are typically labelled as \emph{strong agents}~\cite{intelligentagents-ker10} in the literature on \ac{MAS}\textemdash{}as opposed to \emph{weak agents}, just providing for \emph{computational autonomy}~\cite{artifacts-jaamas17}.
For instance, traditional approaches to \ac{AI} often model decision-making processes around \emph{rational agents}  borrowing concepts and terminology from folk psychology~\cite{folkpsychology-cognition50}. 
\ac{BDI} agents represent a cornerstone of \ac{AI} research, by offering a powerful framework for modelling complex decision-making processes in autonomous systems.
First proposed by \citeap{bratman-compint4}, the \ac{BDI} model introduces a conceptual framework around which  autonomous rational agents based on \emph{beliefs}, \emph{desires}, and \emph{intentions} can be designed and implemented.
The importance of \ac{BDI} agents mostly stems from their ability to provide a structured approach to reasoning, planning, and execution, making them an effective solution for real-world scenarios where autonomous components need to operate in dynamic and uncertain environments. 
For example, \citeap{agentprogramming-jaamas34} argue that \ac{BDI} agents offer robust and flexible behaviour, rapid and modular development, intelligibility, and verifiability.
However, as the complexity of agent environments grows, the need for more sophisticated \textendash{} and, most of all, \emph{faster} \textendash{} decision-making capabilities becomes increasingly critical.

Given the human-like capabilities reached by several \ac{ML} models \textendash{} such as \acp{NN} and \acp{LLM} \textendash{} over a wide variety of perceptual and cognitive tasks \textendash{} e.g., computer vision~\cite{dlobjdetect-tnnls30}, \ac{NLP}~\cite{dlnlpsurvey-tnnls32}, graph processing~\cite{gnn2gnn-uai2022,gnnforifa-eurosp2023}, etc.\ \textendash{}, several researchers have explored the possibility of leveraging \ac{ML} within rational agents~\cite{agentsllm-fcs18}\textemdash{}or, the other way around, of exploiting agents as (encapsulating) abstractions for \ac{ML} technologies in the engineering of intelligent systems~\cite{aose-jaamas9}.
The \ac{ML} component is used to effectively implement one or more perceptual tasks which would be otherwise hard to manage at the \sym{} level typical of the rational agent frameworks.
For example, \citeap{agentslargemodels-aamas2024} promote generative \ac{BDI} architectures where the \ac{BDI} model is coupled with generative \ac{AI} technologies in key steps of the reasoning cycle\textemdash{}such as the management of beliefs, goals, and intentions.

The integration of \ac{ML} techniques into \ac{BDI} agents is becoming an increasingly-popular trend, with dozens of new frameworks being proposed every year~\cite{embodiedagentsllm-iclr2024,recmind-naacl2024,chateda-tcad43,bdinl-aamas2024,voyager-tmlr2024}.
Indeed, integrating \ac{ML} into \ac{BDI} agents allows leveraging on the strengths of both paradigms: the structured reasoning and planning capabilities of \ac{BDI} agents can be augmented with the adaptability and flexibility of \ac{ML} algorithms.
Throughout this manuscript, we refer to agents adopting \ac{ML} for one or more of its cognitive tasks and defined following a \ac{BDI} \textendash{} or, a \ac{BDI}-like \textendash{} architecture as \emph{\mlbdi{} agents}.
Such agents can learn from experience, adapt to new situations, and improve their decision-making capabilities over time, making them even more effective in real-world applications such as robotics, autonomous vehicles, human-computer interactions, and more. 
Moreover, the promotion of \ac{ML} within \ac{BDI} agents allows for incorporating large amounts of data, enabling the agents to learn from complex patterns and relationships that may not be easily captured through traditional rule-based approaches.
In a nutshell, integrating machine learning with \ac{BDI} agents can provide:
\begin{inlinelist}
    \item \emph{improved decision-making}, as \ac{ML} provides the agent with the ability to learn from experience and adapt to new situations;
    \item \emph{increased flexibility}, given that \ac{ML} enables \ac{BDI} agents to operate in environments with high levels of uncertainty and complexity; and
    \item \emph{enhanced autonomy}, \mlbdi{} agents can make decisions without requiring explicit programming.
\end{inlinelist}

While \mlbdi{} agent frameworks are becoming increasingly popular, the methodological and practical exploitation of \ac{ML} techniques during the design and development of \ac{MAS} is yet unclear. 
Indeed, as mentioned by \citeap{agentslargemodels-aamas2024}, the current way of exploiting popular \ac{ML} models for the definition of agent frameworks is leading to a sort of \emph{eliminativism}, where cognitive, mentalistic abstractions such as \ac{BDI} ones are deemed unnecessary, and are increasingly ignored.
Whereas this approach may make the path towards some form of agent intelligence easier or faster, with possibly no consequences over simple \ac{MAS} with few agents, it generally harms the foundations of the software engineering process for intelligent systems~\cite{aose-jaamas9}. 

Therefore, the current state of the art calls for a \emph{fine-grained systematisation} effort, in which the available \mlbdi{} approaches are analysed in depth following the perspective of a rational agent architecture such as the \ac{BDI} one.
To this end, in this paper we survey existing approaches integrating \ac{ML} techniques into rational agents, by observing each work through the lens of the \ac{BDI} architecture, where each basic component of the \ac{BDI} reasoning process is used as a reference module.
Altogether, this also enable us to identify the key research opportunities and open challenges for the effective integration of \ac{ML} into each component of a \ac{BDI} agent. 


\section{Background}\labelsec{background}

\subsection{Belief-Desire-Intention Agents}

The \acf{BDI} framework can be described according to three different perspectives~\cite{pnagent-isda2008}:
\begin{inlinelist}
	\item \emph{philosophical}, based on the work by \citeap{bratman-books1987}, borrowing concepts and terms from folk psychology to view humans as planning agents, where the main concepts are \emph{beliefs} (what the agent knows about the world), \emph{desires} (what the agent wants) and \emph{intentions} (what the agent actually pursues with its actions);
	\item \emph{logical}, mainly based on the work by \citeap{bdilogic-jlc8} on \ac{BDI} logic \textendash{} a multimodal temporal logic with possible world semantics \textendash{}, providing beliefs, goals (desires), and intentions with a precise \emph{logical semantics}; and 
	\item \emph{implementation}, adopted by a large number of technologies and programming frameworks leveraging on the abstract \ac{BDI} model \textendash{} that is, exploiting \ac{BDI} main abstractions (or, some equivalent concepts) as the main agent data structures \textendash{} and conforming to the \ac{BDI} logic for the implementation of the reasoning process.
\end{inlinelist}

Whereas all three the viewpoints are of some relevance here, in the following we mostly focus on \ac{BDI} as the foundation of practical reasoning, designed to simulate human-like reasoning, decision-making, and deliberation processes\textemdash{}and, as such, as the reference architecture for rational agents~\cite{bratman-compint4}.
As one may expect, the \ac{BDI} reasoning process determines the agent behaviour based on \emph{beliefs}, \emph{desires}, and \emph{intentions}.
There, \emph{beliefs} constitute the \emph{epistemic} state of the agent\textemdash{}that is, the agent's knowledge about the world, including itself and other agents, essentially representing what the agent believes to be true.
Beliefs do not necessarily represent the state of the world around the agent, but rather the (possibly partial, possibly inaccurate) view that agent has about its own environment.
\emph{Desires} represent the agent's goals, preferences, and values, depicting the state(s) of the world that the agent aims at achieving.
As such, desires represent the \emph{motivational} state of the agent\textemdash{}that is, what drives its deliberation process and its course of action.
Desires adopted for active pursuit are typically referred to as \emph{intentions}, adding the further restriction that the set of adopted intentions must be consistent (whereas desires need no consistency). 
As such, intentions constitute the \emph{deliberative} state of the agent, representing what the agent has deliberated to actively pursue with its course of action.
From an implementation perspective, intentions identify the agent's plans, and thus determine its subsequent actions.
Pursuing an intention typically means the agent selected a \emph{plan} for execution.
In most \ac{BDI} architectures and implementations, plans are pre-determined (and not dynamically computed by the planning agent) as sequences of actions to be performed by an agent to achieve one or more of its goals, possibly causing the subsequent execution of other plans.
Overall, the \ac{BDI} model uses a deliberation process to reason about the agent's beliefs, desires, and intentions, and execute actions accordingly.

In order to form its beliefs, the agent first receives inputs or percepts about the environment through a \emph{sensing} process, relying on sensors of some sort.
The percepts are then submitted to a \emph{\ac{BRF}}, which supervises the updating process of the agent's beliefs, ensuring that the received percepts do not logically conflict with any existing beliefs.
The \ac{BRF} accepts both the set of percepts and the set of beliefs as input, thus setting a feedback loop.
Afterwards, the agent proceeds by generating a set of desires depending on its updated beliefs about the world and its intentions as well, relying on an \emph{option generation} function. 
Using a \emph{filtering} function, the generated desires are filtered so as to become intentions, checking for realistic and achievable desires depending on the agent's beliefs about the world.
Finally, selected intentions set the agent's forthcoming plans, determining the actions to take \textendash{} through the agent's \emph{actuators} / \emph{effectors} \textendash{} so as to actually change the world and possibly reach the agent's goals.

\subsection{Machine Learning}

In the general \ac{AI} field, \acf{ML} specifically aims at providing machines with the ability to automatically learn from data and past experiences in order to identify patterns and make predictions with minimal human intervention.
As such, \ac{ML} methods enable computers to operate autonomously without explicit programming, aiming at deriving insightful information from large volumes of data by leveraging some optimisation algorithm to identify effective patterns and learn iteratively. 
The optimisation procedure typical of \ac{ML} algorithms uses computational methods to learn directly from data instead of relying on any predetermined equation that may serve as a model.
The optimisation procedure is often called the \emph{learning} or \emph{training} process: it allows the \ac{ML} algorithm to adaptively improve its performance as the amount of available samples grows.

Three major approaches to \ac{ML} exist: namely, \emph{supervised}, \emph{unsupervised}, and \emph{reinforcement} learning.
Each approach is tailored to a well-defined pool of tasks, which may, in turn, be applied in a wide range of use case scenarios.

Supervised learning involves (obviously) supervision, where machines are trained on labelled datasets to predict outputs based on the provided training~\cite{supervisedml-beth51}. 
The labelled dataset specifies the mapping between each input and each output, allowing the optimisation of the machine so as to find a way to approximate an unknown relation.
Here, the optimisation process relies on the definition of a cost function and an optimiser whose aim is to reduce the cost function as much as possible, depending on the given examples.

In unsupervised learning, the machine is optimised using an unlabelled dataset to predict the output, usually relying on some grouping of the unsorted dataset based on the input's similarities, differences, and patterns~\cite{unsupervisedml-cbs2018}.
The learning task consists of finding the best relation for a sample of items, following a given optimality criterion, and intentionally describing the target relation.

\Ac{RL} is a feedback-based process where the machine automatically estimates optimal plans given the reward it receives when it reaches particular goals~\cite{rl-jair4}. 
The machine is iteratively rewarded for each good action and penalised for every wrong move, aiming at maximising the rewards by performing good actions.
There, the rewards constitute the experience, while plans can be described as relations among the possible states of the world, the actions to be performed in those states, and the rewards the agents expect to receive from those actions.

Most notable algorithms include, among the many others, (deep) \acl{NN}, \ac{DT}, \ac{SVM}, random forests, (generalised) linear models, and nearest neighbours.
Since a detailed survey of the available \ac{ML} algorithms is out of the scope of this paper, we refer interested readers to~\cite{cnn-tnnls33,rnn-corr1801.01078,transformers-aiopen3,svmclassification-neucom408}.
Finally, please notice that throughout the remainder of this paper we refer to \ac{ML} algorithms also as \ac{ML} \emph{models}, given the wide adoption of the term in the literature.
Moreover, we also refer to \ac{ML} models relying on purely numerical processing pipelines \textendash{} such as \acl{NN}, where the model is expressed through numerical parameters called \emph{weights} \textendash{} as \emph{\subsym{}} models, thus distinguishing them from \emph{symbolic} approaches, which rely instead on the processing and manipulation of (human-interpretable) symbols.
%

\section{Related Work}\labelsec{related}
%

Several recent surveys explore the integration of \ac{ML} techniques into agent-based systems, with a particular emphasis on multi-agent settings. 
\citeap{MarlChallengesSurvey2021,MarlSurvey2022} provide comprehensive overviews of \marld{}, focusing on the fundamental components of \ac{RL} that enable agents to learn from environmental feedback. 
Complementing this work, \citeap{MarlCommSurvey2024} investigate the communication and interaction mechanisms among agents in \acs{MARL} systems. 
Meanwhile, \citeap{DistributedSwarmSurvey2025} provide a broader distributed perspective examining the role of \ac{ML} in autonomous agent swarms, framing distributed \ac{ML} as a key enabler for large-scale \ac{MAS}.
Adopting a more generalist viewpoint, \citeap{MlAgentTasksSurvey2025} survey the use of \ac{ML} across the agent lifecycle \textendash{} from design to experimentation \textendash{} highlighting the problems \ac{ML} can help addressing within the agent-based modelling processes.

Driven by the increasing capabilities of \acp{LLM}, a significant portion of recent literature has shifted toward the integration of \acp{LLM} into agent architectures. 
\citeap{agentsllm-fcs18} present the first systematic review of \ac{LLM}-based agents, proposing a unified framework for leveraging \acp{LLM} as \subsym{} reasoning modules inside agents.
Similarly, \citeap{LlmMasSurvey2024a,LlmMasSurvey2024b} adopt a holistic perspective, systematising the use of \acp{LLM} in \ac{MAS} and identifying common workflows, infrastructures, and open challenges.
Several surveys focus on specific aspects of \acs{LLM}-based agent design.
\citeap{PlanSurvey2024} present a taxonomy of planning strategies in \acs{LLM} agents, classifying existing approaches across five dimensions: task decomposition, plan selection, external modules, reflection, and memory. 
\citeap{SocialAgentSurvey2025} analyse \ac{LLM}-based agents through the lens of social science and simulation, raising concerns about their realism and the lack of rigorous validation, and arguing that \acp{LLM} may exacerbate long-standing challenges in agent modelling.

The evaluation of \ac{LLM}-based agents has also emerged as a central concern, particularly as they are increasingly deployed in real-world and enterprise contexts.
\citeap{EnterpriseAgentSurvey2025} introduce a two-dimensional taxonomy for evaluating \ac{LLM} agents \textendash{} based on evaluation objectives and processes \textendash{} while also highlighting enterprise-specific concerns such as reliability and compliance. 
Extending this, \citeap{EvaluationAgentsSurvey2025} propose a four-dimensional framework that includes agent capabilities, domain-specific benchmarks, generalist benchmarks, and evaluation methodologies.

Lastly, few other surveys address domain-specific applications, summarising how \acp{LLM} and \ac{ML}-based agents can be applied to chemistry~\cite{LlmAgentChemistrySurvey2024}, web automation~\cite{LlmWebAgentSurvey2025}, and more~\cite{MlAgentOilSurvey2019}. 
Additionally, \citeap{LlmAgentTrustSurvey2025} focus on the critical and often overlooked issue of trust in \ac{LLM}-based agents, summarising emerging attacks and defenses against \ac{LLM} engines.

While these surveys provide valuable insights, they largely follow the prevailing trend of using agent abstractions to support \ac{ML} technologies in the engineering of intelligent systems. 
However, they often neglect cognitive and mentalistic abstractions \textendash{} such as those found in rational agent models like the \ac{BDI} architecture \textendash{} which are essential for designing reliable and trustworthy intelligent agents. 
This lack of grounding in established agent theory weakens the software engineering foundations of intelligent systems.
Conversely, in this survey, we adopt the perspective of the well-established \ac{BDI} rational agent architecture, and frame existing approaches that integrate \ac{ML} techniques into rational agents within this structure. 
By doing so, we aim at providing future \ac{ML}-enhanced agent development with a principled and coherent foundation.
%

\section{Integrating ML into BDI Agents}\labelsec{preliminaries}

Aiming at exploiting \ac{ML} capabilities, several researchers exploit \ac{ML} to inject intelligence into agents~\cite{agentsllm-fcs18}.
A popular trend is to rely on \acp{LLM}~\cite{llmsurvey-techrxiv} and foundation models~\cite{ZhouArxiv2023} to enable information management and reasoning.
From an experimental perspective, those \ac{ML}-enabled agents showcase remarkable performance. 
However, as mentioned by \citeap{agentslargemodels-aamas2024}, the current way of building and exploiting popular \ac{ML} models for the definition of agent frameworks is leading to a sort of \emph{eliminativism}, where higher-level abstractions are deemed unnecessary once lower-level ones are fully understood.
Eliminativism represents a dangerous trend, as it neglects the utility of high-level abstractions, which instead are fundamental to designing agent frameworks correctly.

Therefore, this paper follows and supports \citeauthor{agentslargemodels-aamas2024}'s vision, and aims at eradicating eliminativism by focusing on the \ac{ML} integration in the context of the \ac{BDI} agent framework.
To this end, we provide a systematisation of how current literature integrating \ac{ML} into agents fits into the standard \ac{BDI} conceptual framework at a \emph{fine-grained} level.
There, ``fine-grained'' means that \emph{each} component from the \ac{BDI} architecture is used for the identification of trends and challenges in the integration of \ac{ML} models into rational agents, thus providing for a framework based on a complete set of high-level rational abstractions.
A pictorial representation of the framework resulting from such a fine-grained approach can be observed in \reffig{paper_distribution}.


\begin{figure}[!t]
    \centering
    \resizebox{\textwidth}{!}{

        \begin{tikzpicture}[
            every function_box/.style = {},
            function_box/.style = {rectangle, align=center,on grid, draw=black, ultra thick,
                minimum size = 6mm, every function_box},
            bdi_box/.style = {ellipse, align=center,on grid, draw=black, ultra thick,
                minimum size = 6mm, every function_box},
            environment/.style = {fill=gray!80, function_box, minimum width=2cm, minimum height = 16.5cm},
            sense/.style = {fill=Emerald!40, function_box, minimum width=3cm, minimum height = 2cm,rounded corners=0.5cm},
            brf/.style = {fill=Emerald!40, function_box, minimum width=3cm, minimum height = 2cm,rounded corners=0.5cm},
            belief/.style = {fill=Emerald!40, bdi_box, minimum width=6cm, minimum height = 1.5cm},
            option/.style = {fill=Apricot!40, function_box, minimum width=3cm, minimum height = 1.5cm,rounded corners=0.5cm},
            desire/.style = {fill=Apricot!40, bdi_box, minimum width=3cm, minimum height = 1.5cm},
            filter/.style = {fill=NavyBlue!40, function_box, minimum width=3cm, minimum height = 1.5cm,rounded corners=0.5cm},
            intention/.style = {fill=NavyBlue!40, bdi_box, minimum width=3cm, minimum height = 1.5cm},
            plan/.style = {fill=NavyBlue!40, function_box, minimum width=7cm, minimum height = 1.5cm,rounded corners=0.5cm},
            act/.style = {fill=Mulberry!40, function_box, minimum width=3cm, minimum height = 1.5cm,rounded corners=0.5cm},
            agent/.style = {function_box, minimum width=22cm, minimum height = 16.5cm,rounded corners=0.5cm},
            title/.style={font=\fontsize{10}{10}\color{black},align=center,on grid},
            citer/.style={font=\fontsize{18}{18}\color{black},align=center,on grid},
            ]

            \node[environment] (environment)  at (0,0)  {\rotatebox{90}{\Huge Environment}};
            \node[sense] (sense) at (4,5.5) {\cite{WeiPromas2012}, \cite{SilvaThesis2020}, \cite{SilvaAamas2021},\\ 
                                            \cite{DipaloArxiv2023}, \cite{embodiedagentsllm-iclr2024}, \cite{ZhaoArxiv2025}, \\
                                            \cite{GuanKdd2024}, \cite{Wirelessagent2025}};
            \node[brf] (brf) at (8.5,5.5) {\cite{JianJim2007}, \cite{TekulveIcdl2019}, \cite{bdihealthcare-jms45},\\ 
                                        \cite{TekulveTcds2022}, \cite{IchidaSac2023}, \cite{bdinl-aamas2024},\\ 
                                        \cite{MelloEsa2024}};
            \node[belief] (belief) at (15.25,5.5) {\cite{JianJim2007}, \cite{LeeWsc2008}, \cite{LeeInforms2009}, \cite{LeeTomacs2010}, \cite{KimUoa2015},\\
                                                \cite{ZhuangMatec2018}, \cite{RabinowitzIcml2018}, \cite{LeEmnlp2019}, \cite{JaraCobs2019}, \cite{CuzzolinPsy2020}, \cite{DissingIjcai2020},\\
                                                \cite{MorenoArxiv2021}, \cite{SclarIcml2022}, \cite{HoTcs2022}, \cite{HuangCorl2022}, \cite{reflexion-neurips2023}, \cite{ZhouAiide2023}, \cite{DipaloArxiv2023},\\
                                                \cite{PengAiide2023}, \cite{KumarHci2023}, \cite{SahaArxiv2023}, \cite{ZhuArxiv2023}, \cite{react-iclr2023}, \cite{NascimentoAcsos2023}, \cite{ZhaoPatterns2023},\\
                                                \cite{LianWww2024}, \cite{bdinl-aamas2024}, \cite{QianAcl2024}, \cite{WuAcl2024}, \cite{recmind-naacl2024}, \cite{JinAcl2024}, \cite{embodiedagentsllm-iclr2024},\\
                                                \cite{ZhaoArxiv2025}, \cite{JiangWc2024}, \cite{KannanIros2024}, \cite{NiEml2024}, \cite{generativeagents-2023}, \cite{retroformer-2024}, \cite{metaagents-2023},\\
                                                \cite{metagpt-2024}, \cite{GaoSthree2023}, \cite{Finrobot2024}, \cite{Recagent2025}, \cite{Personalwab2025}, \cite{Wirelessagent2025}, \cite{Codriving2025}};
            \node[option] (option) at (22,5.5) {None};
            \node[desire] (desire) at (15.25, 1.5) {\cite{RabinowitzIcml2018}, \cite{JaraCobs2019}, \cite{CuzzolinPsy2020}, \cite{OguntolaRoman2021}, \cite{KumarHci2023},\\
                                                \cite{embodiedagentsllm-iclr2024}, \cite{KannanIros2024}, \cite{NiEml2024}, \cite{generativeagents-2023}, \cite{metaagents-2023},\\
                                                \cite{metagpt-2024}, \cite{FreringEaai2025}, \cite{chatbdi-2025}, \cite{GaoSthree2023}, \cite{Finrobot2024}};
            \node[filter] (filter) at (7, 0) {\cite{NottinghamIcml2023}, \cite{IchidaSac2023}, \cite{bdinl-aamas2024}};
            \node[intention] (intention) at (15.25, -1.5) {\cite{TahboubJirs2006}, \cite{DiaconescuPloscb2014}, \cite{NguyenRoman2018}, \cite{RabinowitzIcml2018}, \cite{JaraCobs2019}, \cite{CuzzolinPsy2020},\\ 
                                                        \cite{OguntolaRoman2021}, \cite{KumarHci2023}, \cite{embodiedagentsllm-iclr2024}, \cite{KannanIros2024}, \cite{NiEml2024}, \cite{generativeagents-2023},\\
                                                        \cite{metaagents-2023}, \cite{metagpt-2024}, \cite{GaoSthree2023}, \cite{Finrobot2024}, \cite{Codriving2025}};
            \node[plan] (plan) at (15.25, -5.5) {\cite{GuerraClima2004}, \cite{GuerraTas2008}, \cite{SubagdjaIat2008}, \cite{bdilearning-ijats1}, \cite{SinghAtal2010}, \cite{TanNeuro2010}, \cite{SinghIjcai2011}, \cite{TanEswa2011}, \\ 
                                            \cite{ShawSaicsit2015}, \cite{RamirezMates2017}, \cite{SchrodtTcs2017}, \cite{WanAcai2018}, \cite{RabinowitzIcml2018},\cite{BoselloEmas2019}, \cite{SacharnyIas2021}, \cite{PulawskiAcsos2021}, \\
                                            \cite{HoTcs2022}, \cite{chainofthoughts-neurips2022}, \cite{HuangIcml2022}, \cite{HuangCorl2022}, \cite{KumarHci2023}, \cite{ZhuArxiv2023}, \cite{hugginggpt-neurips2023}, \cite{GramopadhyeIros2023}, \cite{HaoEmnlp2023}, \\
                                            \cite{react-iclr2023}, \cite{RanaCorl2023}, \cite{SongIccv2023}, \cite{NottinghamIcml2023}, \cite{DipaloArxiv2023}, \cite{WuArxiv2023}, \cite{XiaEtfa2023}, \cite{HuangArxiv2023}, \\ 
                                            \cite{chateda-tcad43}, \cite{NascimentoAcsos2023}, \cite{ColasPmlr2023}, \cite{swiftsage-neurips2023}, \cite{RuanArxiv2023}, \cite{LianWww2024}, \cite{voyager-tmlr2024}, \cite{bdinl-aamas2024}, \\
                                            \cite{chateda-tcad43}, \cite{recmind-naacl2024}, \cite{embodiedagentsllm-iclr2024}, \cite{GuanKdd2024}, \cite{ZhaoArxiv2025}, \cite{JiangWc2024}, \cite{KannanIros2024}, \\
                                            \cite{NiEml2024}, \cite{generativeagents-2023}, \cite{retroformer-2024},\cite{metaagents-2023}, \cite{tptuv2-2024}, \cite{metagpt-2024}, \cite{SchulzIjpeds2025}, \\
                                            \cite{CiattoEcai2025}, \cite{GaoSthree2023}, \cite{Finrobot2024}, \cite{Personalwab2025}, \cite{Wirelessagent2025}, \cite{Codriving2025}};
            \node[act] (act) at (6, -5.5) {\cite{AmadoIjcnn2018}, \cite{PereiraIjcai2019}, \cite{ZhiNips2020},\\ 
                                        \cite{DipaloArxiv2023}, \cite{WuArxiv2023}};
            \node[agent] (agent) at (13,0) {};

            \draw [-{Latex[length=4mm]},line width=0.75mm] (1,5.5) -- (2.5,5.5); 
            \draw [-{Latex[length=4mm]},line width=0.75mm] (5.5,5.5) -- (7,5.5); 
            \draw [-{Latex[length=4mm]},line width=0.75mm] (10,5.5) -- (10.6,5.5); 
            \draw [-{Latex[length=4mm]},line width=0.75mm] (19.85,5.5) -- (20.5,5.5); 
            \draw [{Latex[length=4mm]}-,line width=0.75mm] (7.5,-5.5) -- (11.35,-5.5); 
            \draw [{Latex[length=4mm]}-,line width=0.75mm] (1,-5.5) -- (4.5,-5.5); 

            \draw [-{Latex[length=4mm]},line width=0.75mm] (15.25,-2.5) -- (15.25,-3.75);

            \draw [-{Latex[length=4mm]},line width=0.75mm, to path={-| (\tikztotarget)}] (21.5,4.75) -- (21.5,1.5) -- (desire);
            \draw [-{Latex[length=4mm]},line width=0.75mm, to path={-| (\tikztotarget)}] (intention) -- (22.5,-1.5) -- (22.5,4.75);
            \draw [-{Latex[length=4mm]},line width=0.75mm, to path={-| (\tikztotarget)}] (desire) -- (7.5,1.5) -- (7.5, 0.75);
            \draw [-{Latex[length=4mm]},line width=0.75mm, to path={-| (\tikztotarget)}] (7.5, -0.75) -- (7.5,-1.5) -- (intention);
            \draw [-{Latex[length=4mm]},line width=0.75mm, to path={-| (\tikztotarget)}] (intention) -- (15.25,0) -- (filter);

            \draw [-{Latex[length=4mm]},line width=0.75mm, to path={-| (\tikztotarget)}] (belief) -- (15.25,3) -- (6.5,3) -- (6.5, 0.75); 
            \draw [-{Latex[length=4mm]},line width=0.75mm, to path={-| (\tikztotarget)}] (belief) -- (15.25,8) -- (9,8) -- (9,6.5); 

            \node[title] at (21.5, -7.5){\Large ML-BDI Agent};

            \node[title] at (4,6.75){Sensing};
            \node[title] at (8.1,6.75){Revision};
            \node[title] at (22, 6.5){Option Generation};
            \node[title] at (6.625, -1.125){Filtering};
            \node[title] at (17, -3.5){Planning};
            \node[title] at (6, -4.5){Acting};

            \draw[color=transparent,postaction={decorate, decoration={text along path, raise=4pt, text align={right,right indent=1.5cm}, text={Beliefs}, reverse path}}] (belief) ellipse (4.6 and 2.0); 
            \draw[color=transparent,postaction={decorate, decoration={text along path, raise=4pt, text align={right,right indent=1.5cm}, text={Desires}, reverse path}}] (desire) ellipse (3.3 and 0.9); 
            \draw[color=transparent,postaction={decorate, decoration={text along path, raise=4pt, text align={right,right indent=0.5cm}, text={Intentions}, reverse path}}] (intention) ellipse (3.95 and 0.9); 

        \end{tikzpicture}

        }
    \caption{Distribution of works available in the literature over the various BDI modules (\ref{item:rq2}). \textcolor{Emerald}{Belief}-related papers are analyzed in \refsec{ml_beliefs}, while \refsec{ml_desires} delves into ML and BDI \textcolor{Apricot}{desires}. ML approaches targeting \textcolor{NavyBlue}{intentions} are studied in \refsec{ml_intentions}. Finally, in \refsec{ml_actions} we investigate the \textcolor{Mulberry}{action}-enabling ML papers.}
    \labelfig{paper_distribution}
\end{figure}
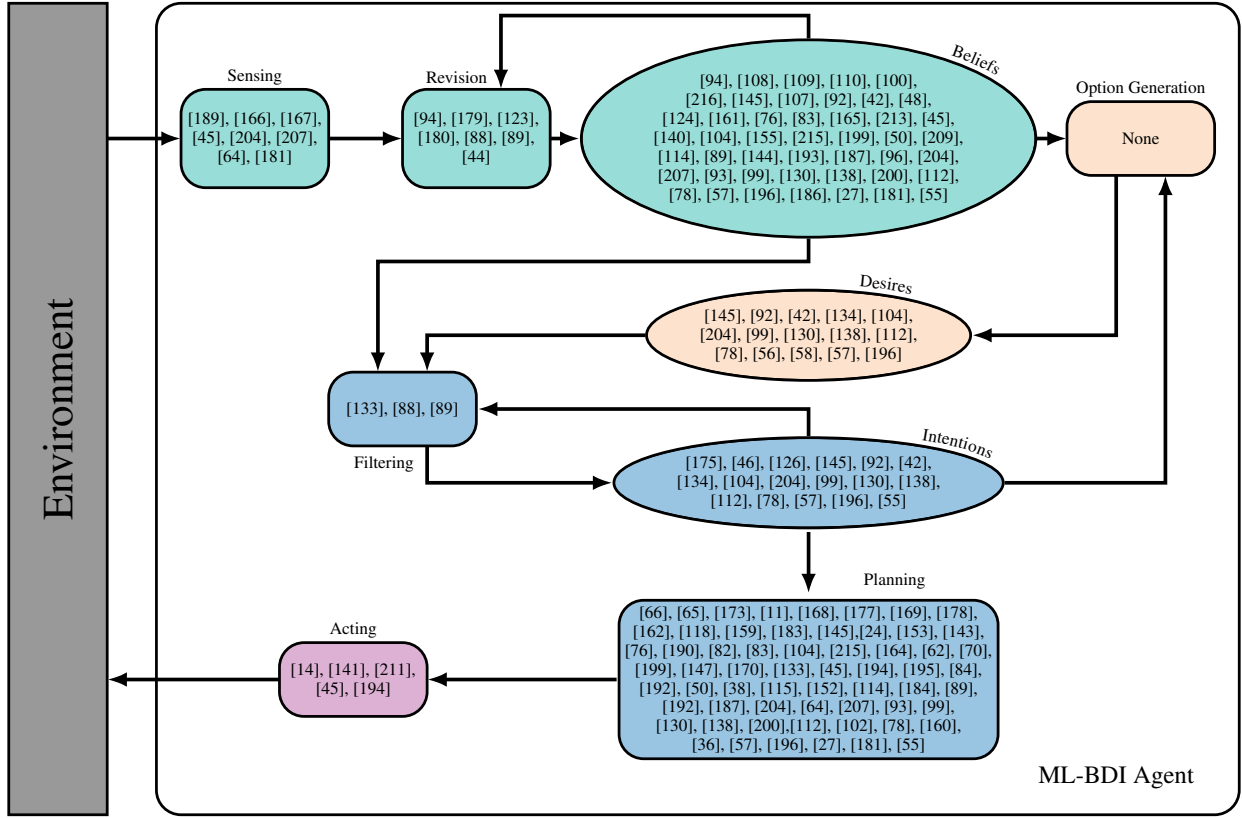

\subsection{Review Methodology}\labelssec{survey}

In this subsection, we delve into the details of how our survey on \mlbdi{} agents is conducted, defining the methodology that we use to identify and analyse the relevant papers available in the literature.
We start by defining six different research questions, namely:
\begin{enumerate}[leftmargin=1.1cm,label=\textbf{RQ\arabic{*}}]
    \item\label{item:rq1} \textendash{} \emph{``How are current approaches integrating \ac{ML} techniques into \mlbdi{} agents?''}
    \item\label{item:rq2} \textendash{} \emph{``For which role are \ac{ML} models used in terms of equivalent \ac{BDI} component?''}
    \item\label{item:rq3} \textendash{} \emph{``Specifically, how are beliefs, desires, and intentions represented in \mlbdi{} agents?''}
    \item\label{item:rq4} \textendash{} \emph{``Which sort of \ac{ML} models are used?''}
    \item\label{item:rq5} \textendash{} \emph{``How is the learning process of \ac{ML} models enabled into \mlbdi{} agents?''}
    \item\label{item:rq6} \textendash{} \emph{``Which and how many \mlbdi{} approaches come with a runnable software implementation?''}
\end{enumerate}
\ref{item:rq1}, \ref{item:rq2}, and~\ref{item:rq3} pivot around the agent perspective, aiming at understanding how the \sym{} reasoning process typical of \ac{BDI} agents is mapped to \subsym{} models in the proposed \mlbdi{} agents.
Meanwhile, \ref{item:rq4} and~\ref{item:rq5} focus on the \ac{ML} perspective, investigating to which extent \mlbdi{} frameworks allow cutting edge \ac{ML} models to be used, and how their optimisation is enabled.
Finally, by answering~\ref{item:rq6} we aim at identifying the state of actual deployability of \mlbdi{} agents.


In order to answer our research questions, we identify several queries to be performed on widely-available bibliographic search engines.
More in detail, we consider the following queries (where $\wedge$ represent a logical AND, $\vee$ is a logical OR):
\begin{itemize}
    \item (`BDI agent' $\vee$ `belief desire intention') $\wedge$ (`machine learning' $\vee$ `learning' $\vee$ `cognitive' $\vee$ `neural networks' $\vee$ `language models' $\vee$ `support vector machine' $\vee$ `decision tree')
    \item (`agent belief' $\vee$ `agent desire' $\vee$ `agent intention') $\wedge$ (`neural networks' $\vee$ `language models' $\vee$ `support vector machine' $\vee$ `decision tree' $\vee$ `learning' $\vee$ `cognitive')
    \item (`intelligent agent' $\vee$ `autonomous agent') $\wedge$ (`neural networks' $\vee$ `language models' $\vee$ `support vector machine' $\vee$ `decision tree' $\vee$ `machine learning' $\vee$ `learning' $\vee$ `cognitive')
    \item (`BDI agent planning') $\wedge$ (`neural networks' $\vee$ `language models' $\vee$ `machine learning' $\vee$ `support vector machine' $\vee$ `decision tree')
\end{itemize}
Here, one should notice that the queries do not focus on \ac{BDI} agents only, targeting instead general intelligent agent frameworks as well. 
This choice allows for the selection of primary works that propose rational agent frameworks, even though not explicitly defined as \ac{BDI}, whose definition can anyway be mapped, either partially or completely, upon the \ac{BDI} architecture.
Reasonably, we consider such approaches as fit to be analysed so as to better understand the state of the integration of \ac{ML} within rational agents.
A possible example is the structure typical of several \ac{LLM}-based intelligent agents~\cite{recmind-naacl2024,embodiedagentsllm-iclr2024,generativeagents-2023}, where the agent's understanding of the world is represented in textual form, which is processed to determine agent's course of action.
While the notion of agent's beliefs is not explicitly mentioned there, that sort of structure can be easily mapped into the belief base of a \ac{BDI} agent, where the agent's knowledge about the world is mapped onto textual beliefs using some \ac{NLP} model.

Back to the methodology used in this paper, as far as bibliographic search engines are concerned, we exploit Google Scholar, Scopus, Springer Link, ACM Digital Library, and DBLP.
For each search engine and query pair, we consider the first five pages of results and inspect the title, abstract, and \textendash{} in case of ambiguity \textendash{} the introduction of each returned paper.
In turn, each paper is assigned to one of the following three disjoint classes:
\begin{inlinelist}
    \item the paper is a \emph{primary work} describing an approach to integrate \ac{ML} models into \ac{BDI} or \ac{BDI}-like agents,
    \item\label{step:secondary} the paper is a \emph{secondary work} surveying some portion of literature about \ac{ML} models and \ac{BDI} \textendash{} or, more generically, rational \textendash{} agents,
    \item the paper is \emph{unrelated} w.r.t.\ to \mlbdi{}, hence it is not relevant for this survey, and is discarded.
\end{inlinelist}
Notably, secondary works selected in step~\ref{step:secondary} are valuable sources of primary works, so they are recursively explored to identify further primary works.
Specifically, the selection process here ultimately led to \allcount{} articles surveyed in this paper.

\subsection{Preliminary Results}\labelssec{preliminary_results}

Before delving into the detailed analysis of each work identified in our survey, we here describe some preliminary results about the literature in its generality.
\reffig{paper_distribution} presents the distribution of the available works over the modules featured by the most common \ac{BDI} frameworks.
We note that some papers may correspond to more than one \ac{BDI} module since they propose agent frameworks where \ac{ML} is used to tackle multiple aspects of the agent's reasoning process.

The papers' distribution looks overall quite unbalanced, with several works focusing on representing the agent's beliefs and planning the agent's actions.
Meanwhile, works targeting more complex aspects of the \ac{BDI} reasoning process that require output verification and/or formal checking \textendash{} such as desire generation, or belief revision \textendash{} are still lacking.
Of the \allcount{} papers surveyed in this article, \belorplancount{} tackle belief representation or planning or both, thus answering~\ref{item:rq2}.
This trend is probably due to the complexity of identifying completely reliable \subsym{} \ac{ML} approaches whose behaviour is verifiable, and at the same time satisfies some strict constraints\textemdash{}possibly expressed \sym{}ally.
Several steps of the \ac{BDI} agent's reasoning process ideally require formal verification, or, at least, a procedure that allows \textendash{} up to some extent \textendash{} checking and, if necessary, patching the agent's updated state, whereas most \ac{ML} approaches rely on black-box models which are cumbersome to open, leaving almost no room for model verification and patching~\cite{AdadiAccess2018,quarrel-extraamas2023}.

The heavy focus on belief representation and planning is also linked to the rising popularity of generative and \ac{NLP} models such as \acp{LLM}, which has provided researchers with tools and platforms to develop intelligent agents able to represent their inner reasoning processes in textual form~\cite{agentsllm-fcs18}.
As a result, several of the belief representation and agent planning papers rely on \acp{LLM} to convert environment inputs into agent's beliefs, and plan their course of action based on them\textemdash{}we refer readers to \refssec{belief_rep} and~\refssec{planning} for more details.
This trend also impacts the distribution of the \ac{ML} models considered in the analysed papers\textemdash{}partially answering~\ref{item:rq4}.
Out of the \allcount{} papers surveyed in this article, \llmcount{} of them consider relying on \ac{LLM} by some means, while only \notnncount{} propose \mlbdi{} agent framework relying on \ac{ML} models that are different from \acp{NN}.

\begin{figure}[!b]
\centering
\begin{minipage}{.315\textwidth}
    \centering
    \includegraphics[width=\linewidth]{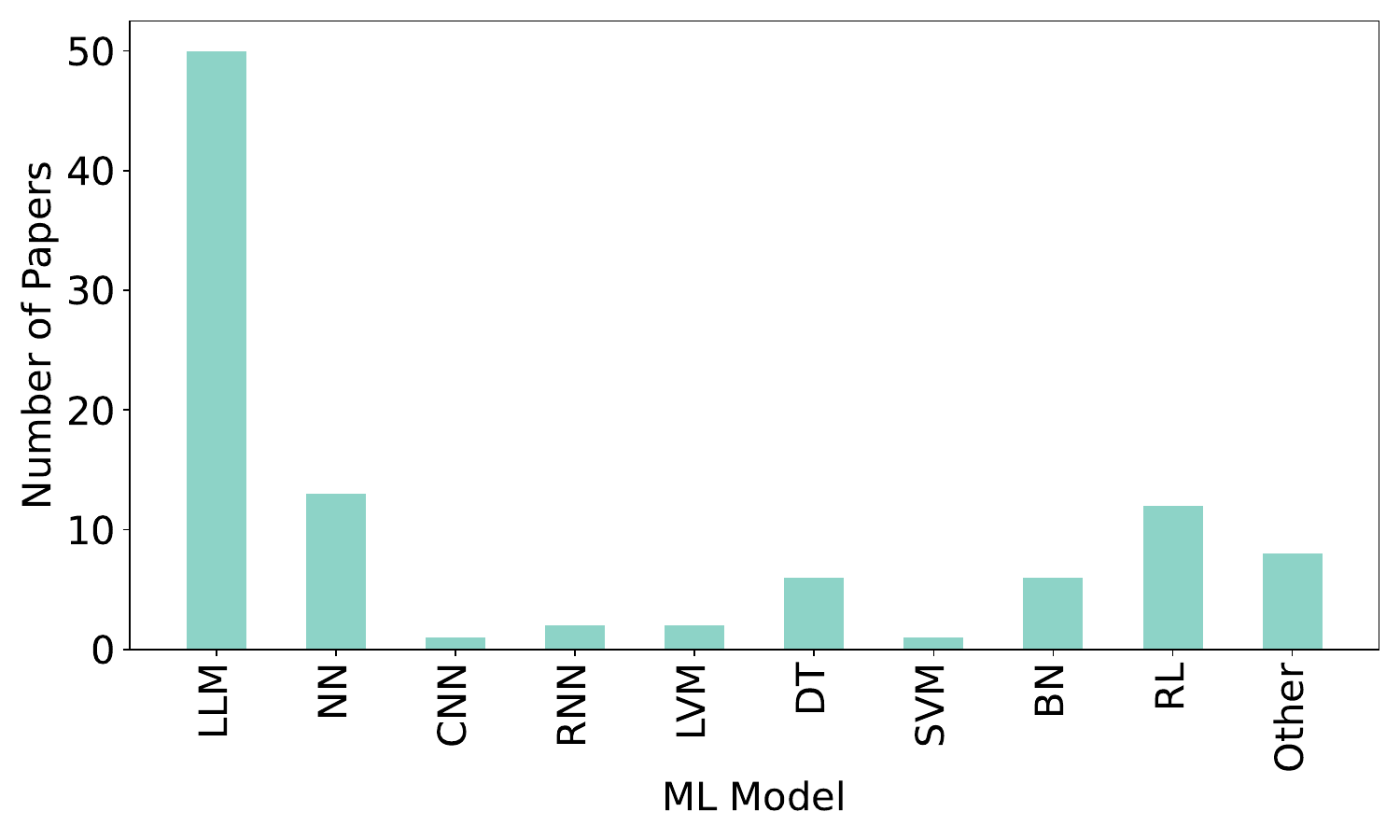}
    \captionof{figure}{Distribution of ML models used in the ML-BDI literature across all years.}
    \labelfig{models_count_plot}
\end{minipage}%
\hspace{0.2cm}
\begin{minipage}{.315\textwidth}
    \centering
    \includegraphics[width=\linewidth]{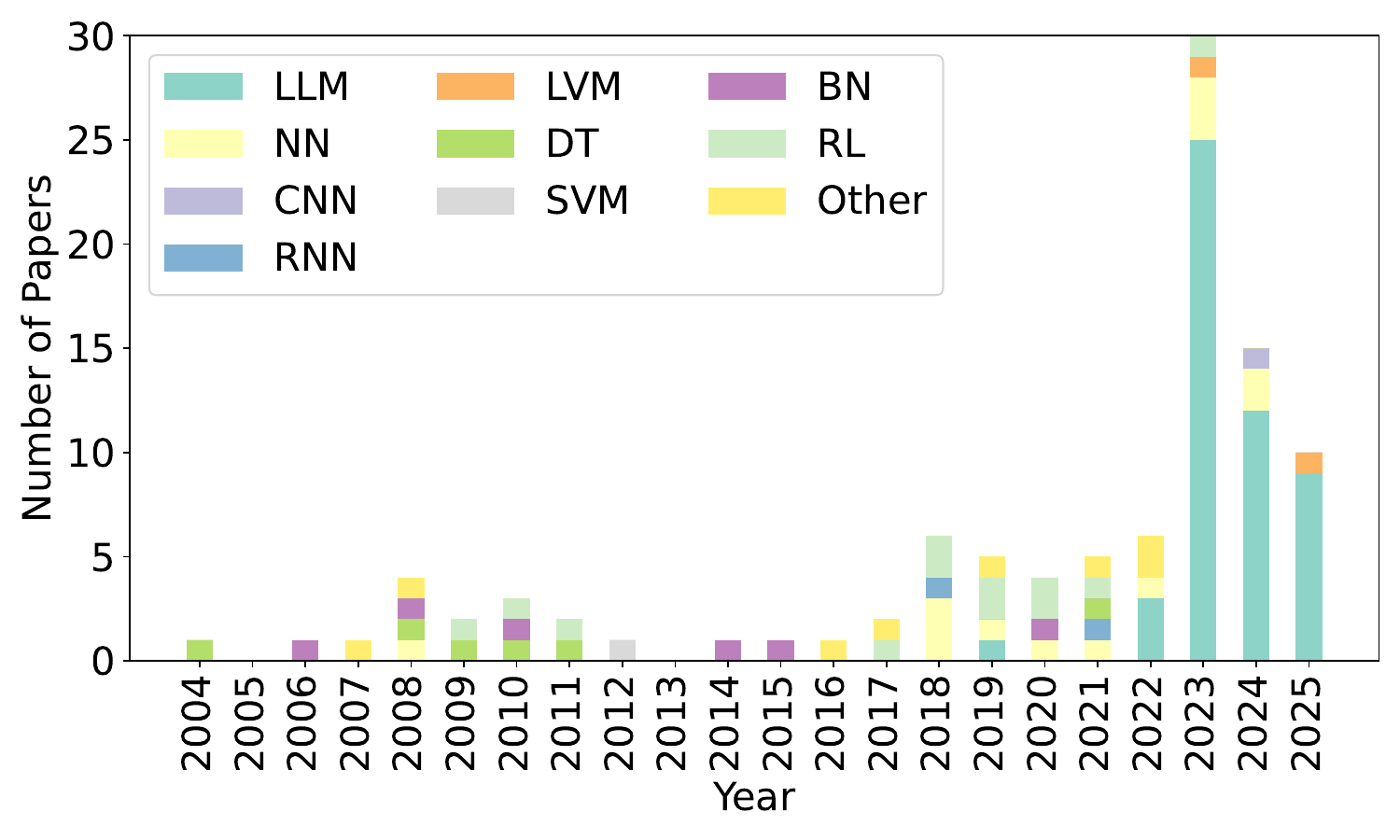}
    \captionof{figure}{Distribution of ML models used in the ML-BDI literature per year.}
    \labelfig{models_count_by_year_plot}
\end{minipage}
\hspace{0.2cm}
\begin{minipage}{.315\textwidth}
    \centering
    \includegraphics[width=\linewidth]{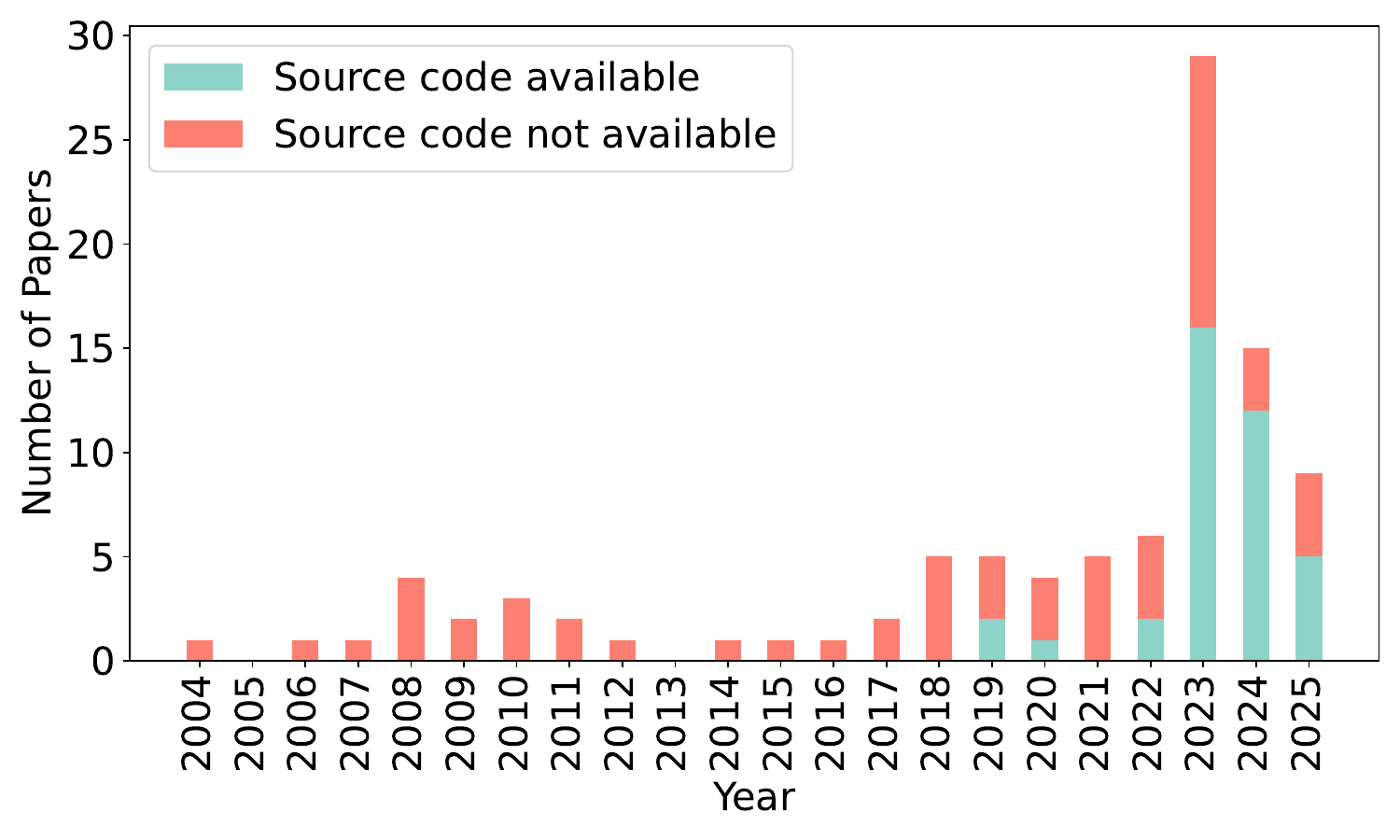}
    \captionof{figure}{Distribution of source code availability of ML-BDI architectures per year.}
    \labelfig{code_count_by_year_plot}
\end{minipage}
\end{figure}

Finally, the popularity of \ac{LLM}-based approaches affects the recency of the papers surveyed, as we here note that \twentyfouronwardcount{} of the considered works were published from 2024 onward, while \twentyonwardcount{} were published after 2020.
The popularity rise highlights the interest of the \ac{ML} research community in developing intelligent agents and the mirrored interest of the agent research community to introduce \ac{ML}-based agent framework to enable \ac{ML} processes into agents.
Moreover, the papers' recency is an implicit proof of the fundamental open research challenges in designing and implementing \mlbdi{} agents.
In this context, we also answer~\ref{item:rq6} by noting that, out of the \allcount{} papers surveyed, only \withcode{} actually come along with publicly-available software implementation\textemdash{}see \reftabfromto{sense-taxonomy}{action-taxonomy} for more details.
Interestingly, \llmwithcode{} out of these \withcode{} target \ac{LLM}-based \ac{BDI}-like agents, showcasing how recent works consider the reproducibility issue more carefully.
\reffigfromto{models_count_plot}{code_count_by_year_plot} show the statistics about \ac{ML} models usage and code availability over the years spanning this systematic review.

\section{Machine Learning and Beliefs}\labelsec{ml_beliefs}

In this section, we present the findings of our literature survey that concern the available approaches integrating \ac{ML} into the \ac{BDI} modules related to the agent's belief perception, processing and representation.
In a nutshell, we find that \ac{ML} techniques are often used to manage and/or improve agent's sensing and beliefs management.
Representing beliefs in textual form and manage/process them using \acp{LLM} represent a popular trend to enable complex \textendash{} although possibly unreliable \textendash{} reasoning about the environment and other agents' states.
We analyse in detail each paper presenting its workflow (thus answering~\ref{item:rq1}) and checking
\begin{inlinelist}
    \item the \ac{ML} model used by each approach (\ref{item:rq4});
    \item the belief structure \textendash{} i.e., how are beliefs represented in the agent \textendash{} answering~\ref{item:rq3};
    \item if the proposed approach supports online learning of the \ac{ML} model inside the agent framework (\ref{item:rq5}); and
    \item if the proposed approach comes with a publicly-available implementation (\ref{item:rq6}).
\end{inlinelist}
%

\subsection{Sensing}\labelssec{belief_sens}

In \ac{BDI} agents, the primary process for acquiring environmental information is sensing/perception~\cite{StabileBracis2015}. 
Apart from specific epistemic actions, this process is typically passive and independent of the agent's internal state, constructing beliefs directly from sensor outputs. 
Here, the term \emph{sensor} is used broadly to refer to any component that enables the agent to sense its environment and form beliefs about its current state.
\ac{ML}-based solutions for intelligent sensing are becoming increasingly common.
In computer vision, for instance, \acp{NN} are widely used to identify objects in the environment. 
Numerous \ac{NN}-based approaches have been proposed for object detection~\cite{dlobjdetect-tnnls30}, pose estimation~\cite{ZhengCsur2024}, and more~\cite{EstevaNature2021}. 
However, we consider surveying all \ac{ML}-based sensory methods to be beyond the scope of this paper, as there exist several surveys already cover \ac{ML} models for object sensing~\cite{ZhangArxiv2023Remote}, intelligent sensor design~\cite{BallardNature2021}, chemical sensors~\cite{ChuSaabc2021} and more. 
We refer interested readers to those works and, instead, focus on papers that explicitly integrate \ac{ML}-based sensing into agent systems.
Our review identifies very few such works, likely due to the \ac{ML} community's focus on general-purpose solutions and the historical separation between autonomous agents and \ac{ML} research.

\citeap{WeiPromas2012} propose a cognitive robot control architecture that uses \acp{SVM}~\cite{svmclassification-neucom408} for object identification.
Their system integrates low-level \subsym{} control \textendash{} via OpenCV\footnote{\url{https://opencv.org}} functions \textendash{} with high-level \sym{} reasoning through the GOAL agent programming language~\cite{HindriksAose2014}.
Meanwhile, in~\cite{SilvaThesis2020,SilvaAamas2021}, the authors focus on the issue of passive sensing that characterises common agent systems having partial perception of the environment.
The authors propose active perception by adjusting camera positions in \ac{ML}-enabled vision systems, integrating this mechanism into the \ac{BDI} reasoning cycle.
\citeap{GuanKdd2024} propose a smart assistant agent  for web surfing, where a YOLOX-based~\cite{Yolox} object detection module clusters webpage components to extract relevant text.
Such information is then processed by an \ac{LLM} engine to optimise subsequent planning steps.
More advanced approaches use language and vision foundation models to design frameworks where language serves as the core reasoning tool.
For example, \cite{DipaloArxiv2023} uses vision models to identify objects and their spatial relationships, while \citeap{ZhaoArxiv2025} apply similar techniques to intelligent aerial agents, focusing on edge deployment of these models for sensing and planning.
In~\cite{embodiedagentsllm-iclr2024}, the authors incorporate a perception module into their agent architecture using Mask-RC\ac{NN}~\cite{HeIccv2017} to predict segmentation masks from RGB images and build a 3D point cloud.
This enables the extraction of high-level information such as object states and the construction of a local semantic map.
Lastly, \acp{LLM} can also function as sensing modules in agent architectures~\cite{Wirelessagent2025} processing textual input from human-agent interactions and storing extracted beliefs in textual form.

\reftab{sense-taxonomy} briefly summarises the main features of the analysed works.

\begin{mybox}
\phantomsection
{{\bf \faSearch\ \textit{Quick Takeaway:}\label{take:sense}} ML-based sensing in agents is underexplored; few frameworks integrating vision or language models into agent perception, mostly targeting object identification.}
\end{mybox}


\newcommand{\sensetabhead}{
	\textbf{\#} & \textbf{Method} & \textbf{Year} & \textbf{ML} (\ref{item:rq4}) & \textbf{On. Learn.} (\ref{item:rq5}) & \textbf{Belief} (\ref{item:rq3}) & \textbf{Tech.} (\ref{item:rq6}) \\
	\hline\hline
}

\newcounter{SenseMethod}
\setcounter{SenseMethod}{1}
\newcommand{\newSenseMethodIndex}{\theSenseMethod\stepcounter{SenseMethod}}

\begin{scriptsize}

\begin{longtable}{c|p{3.5cm}|c|c|c|c|c}
    \caption{
        Summary of \ac{ML}-enabled sensing in \mlbdi{} agents. Legend: G=\emph{goal}, U=\emph{unspecified}, J=\emph{jason}, LVM=\emph{large vision model}, LLM=\emph{large language model}, V=\emph{vectorial}, CNN=\emph{convolutional neural network}, L=\emph{language}.
    }
    \labeltab{sense-taxonomy}\\        
    \sensetabhead
    \endfirsthead
    \caption[]{Summary of \ac{ML}-enabled sensing in \mlbdi{} agents (continued).}\\
    \sensetabhead
    \endhead
    \endlastfoot
    \newSenseMethodIndex & \citesp{WeiPromas2012} & SVM & \xmark & G & \unavailable
    \\\hline
    \newSenseMethodIndex & \citesp{SilvaThesis2020} & U & \xmark & J & \unavailable
    \\\hline
    \newSenseMethodIndex & \citesp{SilvaAamas2021} & U & \xmark & J & \unavailable
    \\\hline
    \newSenseMethodIndex & \citesp{DipaloArxiv2023} & LVM & \cmark & V & \unavailable
    \\\hline
    \newSenseMethodIndex & \citesp{embodiedagentsllm-iclr2024} & CNN & \xmark & L & \tiny{\url{https://tinyurl.com/29kecyhe}} 
    \\\hdashline
    \newSenseMethodIndex & \citesp{GuanKdd2024} & CNN & \xmark & L & \unavailable
    \\\hline
    \newSenseMethodIndex & \citesp{ZhaoArxiv2025} & LVM & \cmark & V & \unavailable
    \\\hdashline
    \newSenseMethodIndex & \citesp{Wirelessagent2025} & LLM & \xmark & L & \tiny{\url{https://tinyurl.com/2bbt2ykd}} 
    \\\hline\hline

\end{longtable}

\end{scriptsize}

\subsubsection*{\textbf{Opportunities and future directions}}
Given the limited number of approaches targeting \ac{ML}-enabled sensing in \ac{BDI} agents, and the growing emphasis on \ac{ML} within the \ac{AI} community, we anticipate increased research interest in this area.
Currently, frameworks that support the integration of advanced \ac{ML}-based sensing into popular agent technologies \textendash{} such as \jason{}~\cite{jasonbook2007}, \agentspeak{}~\cite{RaoMaamaw1996}, \jacamo{}~\cite{jacamo-scp78}, and others~\cite{HindriksAose2014} \textendash{} are entirely missing.
We therefore emphasize the need to develop such integration frameworks to enable the definition and deployment of \mlbdi{} agents in real-world environments.
When incorporating \ac{ML} models for environmental sensing, it is essential to account for the probabilistic nature of their predictions.
\ac{ML} outputs must be processed and converted into relevant, reliable beliefs.
However, this conversion is non-trivial, as \subsym{} sensing outputs are inherently uncertain\textemdash{}\acp{NN} can be unreliable or overly confident~\cite{SzegedyIclr2013,NguyenCvpr2015}.
Future research should explore how to translate \ac{ML} sensing outputs into beliefs, potentially using probabilistic or para-consistent representations. 
Promising directions include probabilistic logic~\cite{RiguzziDlp2018}, para-consistent logic~\cite{VilladsenPcl2002,AngelottiIccima2001} and probabilistic agent frameworks~\cite{DixTocl2000,NguyenIjcai2019}.

Additionally, the issue of partial perception in real-world environments remains critical. 
Building on the assumptions in~\cite{SilvaAamas2021}, we advocate for \ac{ML}-based solutions that support \emph{belief-driven} sensing.
An agent's belief set can guide the sensing process by identifying aspects of the environment that require closer attention.
For instance, \ac{ML} models could predict the variability of environmental parameters, allowing agents to prioritise sensing of highly dynamic features.
Stable variables \textendash{} such as indoor temperature for smart robots or road surface characteristics for autonomous vehicles \textendash{} may be monitored less frequently, while volatile ones require continuous observation.
Yet, identifying relevant parameters and their variability in complex environments is challenging and \ac{ML} techniques may help automate this process.
Similarly, integrating \ac{ML} models into the \ac{BDI} sensing process could enable \emph{goal-driven} sensing.
In applications like autonomous driving, it is beneficial for agents to focus on environmental aspects most relevant to their goals.
\ac{ML} models could be optimised to identify and prioritise such parameters.
In this context, \acf{NeSy}~\cite{GarcezAir2023} and \acf{SKI}~\cite{skeislr-csur56} offer promising avenues for enabling \emph{belief-} and \emph{goal-driven} sensing.
These approaches allow \ac{ML} models to be constrained by the agent's beliefs or goals, steering the sensing process toward specific environmental features.
The growing interest in integrating \ac{NeSy} and \ac{SKI} into agent systems~\cite{skiqos-jaamas37,skenlp-jaamas38,MelloEsa2024} highlights a promising trend in this research direction.

\subsection{Belief Revision}\labelssec{belief_rev}

Belief revision is a fundamental component of \ac{BDI} agents, required when the agent receives conflicting information from different sources or data inconsistent with its current beliefs. 
In practice, revising beliefs to restore consistency is non-trivial due to the complexity of detecting, identifying, and resolving conflicts. 
\ac{ML}-enabled frameworks for managing such conflicts present a promising opportunity for \mlbdi{} research.
However, despite the importance of this problem, our systematisation identified only \brfcount{} works addressing this module.
Most focus on converting sensing outputs into beliefs, often lacking a full characterisation of belief revision functions and overlooking conflict resolution.

\citeap{JianJim2007} explore a multi-agent setting using ART \textendash{} an instance of \ac{SONN} \textendash{} to model agent beliefs.
Agents receive information about others' skills via \ac{SONN} embeddings and apply ART to convert these into beliefs.
Advertised skills form initial beliefs, while actual capabilities are learned through interaction, updating the embeddings.
In~\cite{TekulveIcdl2019,TekulveTcds2022}, neural dynamic networks enable a robotic agent to form, activate, and reject beliefs in a simulated task.
Activated concept nodes are linked to belief nodes via a reward-modulated Hebbian learning rule, with belief activation following a neural match operation akin to ART's resonance principle.

\citeap{bdihealthcare-jms45} integrate a \ac{ML}-based cognitive service into a \jason{}~\cite{jasonbook2007} (\ac{BDI}) agent.
The service suggests actions based on a decision tree prediction, adding beliefs to the agent.
The agent then decides whether to follow the suggestion or rely on expert-encoded knowledge.
In~\cite{IchidaSac2023,bdinl-aamas2024}, a conversational \ac{BDI} agent uses \acp{NN}-based \ac{NLP}~\cite{dlnlpsurvey-tnnls32} to interpret human prompts and convert them into beliefs about user requests, dynamically revising prior beliefs.
Finally, \citeap{MelloEsa2024} integrate \ac{NN} into the \ac{BDI} pipeline \textendash{} implemented using the Sigon framework~\cite{GelaimEsa2019} \textendash{} to automate the sensing phase. 
The \ac{NN} processes environmental perceptions, and its output is used as a belief to precondition plan selection.

\reftab{brf-taxonomy} briefly summarises the main features of the analysed works.

\begin{mybox}
\phantomsection
{{\bf \faSearch\ \textit{Quick Takeaway:}\label{take:revise}} ML-enabled belief revision remains rare; existing works focus on belief updates rather than conflict resolution, leaving room for improvement.}
\end{mybox}


\newcommand{\brftabhead}{
	\textbf{\#} & \textbf{Method} & \textbf{Year} & \textbf{ML} (\ref{item:rq4}) & \textbf{On. Learn.} (\ref{item:rq5}) & \textbf{Belief} (\ref{item:rq3}) & \textbf{Tech.} (\ref{item:rq6}) \\
	\hline\hline
}

\newcounter{BrfMethod}
\setcounter{BrfMethod}{1}
\newcommand{\newBrfMethodIndex}{\theBrfMethod\stepcounter{BrfMethod}}

\begin{scriptsize}

\begin{longtable}{c|p{3.5cm}|c|c|c|c|c}
    \caption{
        Summary of \ac{ML}-enabled belief revision in \mlbdi{} agents. Legend: V=\emph{vectorial}, NDF=\emph{neural dynamic fields}, SL=\emph{signal-like beliefs}, J=\emph{jason}, PAS=\emph{python agent-speak}, L=\emph{language-based}, S=\emph{sigon}.
    }
    \labeltab{brf-taxonomy}\\        
    \brftabhead
    \endfirsthead
    \caption[]{Summary of \ac{ML}-enabled belief revision in \mlbdi{} agents (continued).}\\
    \brftabhead
    \endhead
    \endlastfoot
    \newBrfMethodIndex & \citesp{JianJim2007} & SONN & \cmark & V & \unavailable
    \\\hline
    \newBrfMethodIndex & \citesp{TekulveIcdl2019} & NDF & \cmark & SL & \unavailable
    \\\hline
    \newBrfMethodIndex & \citesp{bdihealthcare-jms45} & DT & \xmark & J & \unavailable
    \\\hline
    \newBrfMethodIndex & \citesp{TekulveTcds2022} & NDF & \cmark & SL & \unavailable
    \\\hline
    \newBrfMethodIndex & \citesp{IchidaSac2023} & NN & \xmark & PAS & \unavailable
    \\\hline
    \newBrfMethodIndex & \citesp{bdinl-aamas2024} & NN & \xmark & L & \tiny{\url{https://tinyurl.com/27uazxbb}} 
    \\\hdashline
    \newBrfMethodIndex & \citesp{MelloEsa2024} & NN & \xmark & S & \unavailable
    \\\hline\hline

\end{longtable}

\end{scriptsize}

\subsubsection*{\textbf{Opportunities and future directions}}

Our analysis reveals that \ac{ML}-enabled belief revision functions remain vastly underexplored, with only a few existing approaches despite numerous open challenges and opportunities. 
\ac{ML} models can assist in rapidly identifying conflicting beliefs or generating alternative ones. 
Given the ability of \acp{NN} to learn complex patterns from data, future research may leverage \subsym{} processing to predict when beliefs or belief sets conflict.
This capability could significantly accelerate the initial phase of belief revision and improve scalability, a known challenge in belief revision~\cite{HerzigKi2017}.
To support this, \ac{ML} models would require online learning mechanisms to periodically update the \ac{ML} model, based on the quality of revised beliefs.
We envision \mlbdi{} agents where belief revision relies on fast \ac{ML}-based conflict detection, complemented by \sym{} tools that verify the integrity of revised beliefs.
This hybrid setup resembles \neusym{} models, which combine \sym{} reasoning with \subsym{} learning to create trustworthy frameworks~\cite{SarkerArxiv2021,skeislr-csur56}.
\Neusym{} models are considered more reliable by design, as they incorporate verifiable symbolic components into the learning process~\cite{keynote-fedcsis2023}.
Moreover, \ac{ML} techniques may also serve as accelerators for \sym{} conflict identification within the agent's belief base, rather than replacing \sym{} reasoning entirely, as recent research has explored \ac{ML}-based acceleration of \sym{} frameworks across various domains~\cite{ChenEcoop2018,HeCcs2021}.

Once conflicting beliefs are identified, the agent must revise its belief set effectively.
Generative \ac{ML} approaches~\cite{CaoTkde2024,ZhangArxiv2023_1,ChangTist2024} offer promising tools for this task.
These models are increasingly popular across domains~\cite{llmsurvey-techrxiv}, proving how \subsym{} systems can reliably generate structured information.
In \ac{BDI} agents, generative \ac{ML} could be used either to
\begin{inlinelist}
    \item optimise an end-to-end belief revision/update pipeline, or
    \item work alongside \sym{} conflict checkers.
\end{inlinelist}
A key challenge here is enabling online optimisation of generative models, which typically require large datasets and substantial computational resources~\cite{BenderFaact2021,BaiArxiv2024}.

Finally, a promising direction is belief compression\textemdash{}identifying smaller, equivalent sets of beliefs to streamline revision.
\ac{ML} models could offer efficient solutions to this problem.
Existing knowledge graph compression techniques~\cite{SafaviIcdm2019,HwangEmnlp2023} provide a useful starting point for exploring belief summarisation to support faster and more effective revision.

\subsection{Belief Representation}\labelssec{belief_rep}

In \ac{BDI} agents, beliefs represent the informational state of the agent, defining the agent's understanding of the world.
Belief representation differs depending on the framework considered, but is usually expressed under \sym{} form.
The usage of \ac{ML}-enabled framework for belief representation represents a relevant opportunity for the \mlbdi{} integration literature as it may, for example, allow for the definition of efficient \subsym{} generation of new valid beliefs as well as the identification of common beliefs between agents.
We identify \beliefcount{} available works falling into this \ac{BDI} module, which can be broadly categorised into three sets of approaches, namely:
\begin{description}
    \item[\Subsym{} belief modelling] in which \ac{ML} models are used to implicitly represent the agent's beliefs or knowledge via the model parameters.
    \item[Belief update] in which \ac{ML} models are used to dynamically update the agent's beliefs, depending on the environment feedback and/or the plans' outcome.
    \item[Knowledge enrichment] in which \ac{ML} models infer novel beliefs about the environment or other agents' state.
\end{description}
In the rest of this section, we delve into the details of each category---with \reftab{belief-taxonomy} summarising our findings for each paper, and~\reffigfromto{belief_aim_plot}{belief_learn_plot} overviewing the papers distribution.

\subsubsection{\Subsym{} belief modelling}
The implicit representation of an agent's belief using \subsym{} models such as \acp{NN} is becoming an increasingly popular solution to integrate \ac{ML} into agents.
While the first proposal for \subsym{} belief modelling dates back to \citeap{JianJim2007}, this approach has gained traction recently due to the impact of \acp{LLM} on the \ac{AI} community.
Given the \ac{LLM}'s ability to encode training data into internal parameters and enable complex reasoning-like behaviours, researchers are increasingly accepting the idea of representing agent knowledge within language model parameters\textemdash{}also known as parametric knowledge~\cite{PanTgdk2023}.

\citeap{JianJim2007} first proposed modelling \ac{BDI} agent beliefs using ART\textemdash{}a form of \ac{SONN}.
Beliefs are mapped into \ac{SONN} embeddings to advertise skills and form adaptive beliefs about other agents.
The belief model is a fuzzy \ac{NN} that takes actor advertisements as initial beliefs and learns actual capabilities through interaction.
While innovative, using traditional \acp{NN} for explicit belief representation is cumbersome due to the difficulty of extracting knowledge from black-box models.
Current approaches instead rely on \acp{LLM} to extract and process implicit beliefs via model prompting~\cite{SahooArxiv2024}.
For instance, Reflexion~\cite{reflexion-neurips2023} defines an agent that stores feedback in natural language and reflects on it, forming an episodic memory that acts as its belief base.
Similarly, RecAI~\cite{LianWww2024} uses prompt engineering to enhance domain-specific capabilities of the \ac{LLM}, implicitly embedding domain beliefs.
In~\cite{NascimentoAcsos2023}, each agent in a multi-agent framework uses an \ac{LLM} to analyse inputs from sensors or other agents and plan accordingly, embedding beliefs about the environment and peers.
MechAgents~\cite{NiEml2024}, Smart-LLM~\cite{KannanIros2024}, and FinRobot~\cite{Finrobot2024} also rely on coordinator agents using \acp{LLM} to reason about other agents via implicitly constructed beliefs.

\Acp{LVM} offer another route for embedding agent beliefs using \ac{ML}.
For example, \cite{DipaloArxiv2023,ZhaoArxiv2025} use a \ac{LVM} for sensing, belief construction, and planning, with the language model serving as a plan generator.

\ac{ToM} modelling within belief modules is another emerging area where \acp{LLM} are gaining popularity.
\ac{ToM} allows agents to infer beliefs about others' knowledge and states, improving interaction.
\citeap{LeEmnlp2019} first tackled \ac{ToM} modelling via question answering, highlighting limitations of \ac{ML} models in this domain.
The COKE dataset~\cite{WuAcl2024} was introduced to optimize \acp{LLM} for extracting \ac{ToM} models from text.
\citeap{SahaArxiv2023} explore a teacher-student setting, where the teacher agent uses a \ac{ToM} model \textendash{} implemented via an \ac{LLM} \textendash{} to tailor explanations to the student's beliefs.
In~\cite{JinAcl2024}, the authors test \ac{LLM} capabilities in answering \ac{ToM}-related questions, defining a \subsym{} embedding of \ac{ToM} for use in \ac{BDI} agents.

Despite their popularity, \acp{LLM} are not the only viable solution for \subsym{} belief representation.
\citeap{SclarIcml2022} propose a symmetric \ac{ToM} model where agents can speak, listen, observe, and move freely.
They use standard \ac{NN} and \ac{RL} approaches to tackle this more complex \ac{ToM} problem, showing that agents with \ac{ToM} outperform standard ones.
Overall, these approaches rely on vector or language-based belief representations and heavily depend on complex \ac{ML} models for reasoning.
Few works support online learning~\cite{SclarIcml2022}, either by training the \ac{LLM} directly or using \ac{LLM}-informed \ac{RL}, while others treat the \ac{LLM} as a static processor~\cite{reflexion-neurips2023,LianWww2024,NascimentoAcsos2023}.
Notably, several \ac{LLM}-based approaches include viable implementations~\cite{LeEmnlp2019,SclarIcml2022,WuAcl2024,JinAcl2024}.

\subsubsection{Belief update}

Preliminary approaches in this category model agent beliefs using \acp{BN}, enabling dynamic updates by modifying conditional dependencies and transition probabilities.
For instance, \citeap{LeeWsc2008} and~\citeap{LeeTomacs2010} use a \ac{BN} as a \emph{perceptual processor} within the belief module, inferring attribute values \textendash{} i.e., beliefs \textendash{} from environmental information, akin to human reasoning.
In~\cite{LeeInforms2009}, the same authors extend this by combining bayesian belief networks with a \ac{RL} model to update beliefs based on environmental feedback and plan outcomes.
Similarly, \citeap{KimUoa2015} propose updating the dataset in the perception module when sufficient new observations are available, thereby updating the \ac{BN} representing the agent's beliefs.
Although not using \acp{BN}, the framework in~\cite{ZhuangMatec2018} also fits this category, employing binary neural networks to distinguish between achievable and unachievable beliefs, mapping knowledge into numerical vectors and pruning the belief base accordingly.

More recently, \ac{LLM}-based belief update approaches have emerged, building on the reasoning capabilities of \acp{LLM}.
These methods typically store agent beliefs or environmental knowledge in textual form, prompting \acp{LLM} to identify necessary updates.
For example, the \emph{Inner Monologue} framework~\cite{HuangCorl2022} introduces grounded closed-loop feedback for robot planning, where textual feedback from the environment is iteratively used to update the agent's belief base.
In~\cite{ZhuArxiv2023}, an agent uses a \ac{LLM} to reason over a textual knowledge base to generate reference plans, select the optimal one, and update the knowledge based on action outcomes.
The \emph{ReAct} framework~\cite{react-iclr2023} prompts \acp{LLM} to produce reasoning traces and actions, incorporating external feedback to refine beliefs.
\emph{Retroformer}~\cite{retroformer-2024} builds on \emph{ReAct} by incorporating environmental rewards to fine-tune the planning \ac{LLM}.
\citeap{recmind-naacl2024} propose \emph{RecMind}, where a \ac{LLM} processes externally stored textual beliefs using an algorithm extending the Tree-of-Thoughts approach~\cite{treeofthoughts-neurips2023}.
Memory modules are also widely used to store and process agent experience \textendash{} via a \ac{LLM} \textendash{} to update belief content, as seen in~\cite{embodiedagentsllm-iclr2024,generativeagents-2023,metaagents-2023,metagpt-2024,Personalwab2025}.
These approaches vary in memory module structure, update mechanisms, and application domains\textemdash{}e.g., social networks~\cite{GaoSthree2023}, recommendation systems~\cite{Recagent2025}, telecommunications~\cite{JiangWc2024,Wirelessagent2025}, and autonomous driving~\cite{Codriving2025}.

As with the \emph{\subsym{} belief representation} class, \ac{ToM} modelling using \ac{ML} is a prominent subcategory.
\citeap{RabinowitzIcml2018} and~\citeap{KumarHci2023} propose neural \ac{ToM} models, representing agent knowledge, state, and plans via three \ac{NN} modules.
The first module updates the agent's observations and world knowledge \textendash{} i.e., beliefs \textendash{} using meta-learning~\cite{HuismanAir2021}.
\citeap{JaraCobs2019} and~\citeap{CuzzolinPsy2020} apply \ac{RL} to update \ac{ToM} models based on feedback about peer states.
In human-computer interaction, \cite{DissingIjcai2020} use \acp{NN} to convert sensory input into epistemic logic representing beliefs about human \ac{ToM}.
\citeap{HoTcs2022} and~\citeap{ZhaoPatterns2023} explore less conventional \ac{ML} approaches \textendash{} causal learning and spiking neural networks \textendash{} to model iterative \ac{ToM} updates.

Overall, belief update approaches typically rely on either \acp{BN} and bayesian representations, or foundational models like \acp{LLM} to process textual beliefs.
Vector embeddings are also used when applying \ac{RL} or standard \acp{NN}.
Few frameworks support online learning, either via \ac{RL} or fine-tuning the \ac{LLM} that processes beliefs.
We argue that enabling online learning is crucial for effective belief modelling, as agent feedback should be continuously evaluated and used to adapt behaviour.
Finally, only a small number of approaches in this category offer publicly available source code, limiting reproducibility and practical deployment.
%

\subsubsection{Knowledge enrichment}

The \emph{knowledge enrichment} category of approaches is the most underrepresented class among the three identified in our literature analysis.
Only five papers fall into this category\textemdash{}the majority of which rely on \ac{LLM} to expand the agent's knowledge by incorporating other agents' beliefs or by accumulating novel beliefs under a specific representation.
For example, \citeap{QianAcl2024} propose a multi-agent framework for software development, where different agents share common knowledge and beliefs through a conversation history processed via \ac{LLM}.
Similarly, the authors in~\cite{ZhouAiide2023} and~\cite{PengAiide2023} present two distinct game agents, both of which accumulate environmental knowledge in the form of a knowledge graph and generate new beliefs through conversations with non-player characters using \ac{LLM}.
\citeap{bdinl-aamas2024} introduce an entirely text-based agent framework in which beliefs, plans, and intentions are represented in natural language and processed using \ac{NN} modules, enabling textual belief generation via \ac{NN}.
Finally, only one approach does not rely on \ac{LLM} or text-based belief representation~\cite{MorenoArxiv2021}: in this case, \ac{RNN} are used to extend the agent's beliefs about other agents' beliefs, thereby enriching knowledge through \ac{RNN}-based \ac{ToM}.

\begin{mybox}
\phantomsection
{{\bf \faSearch\ \textit{Quick Takeaway:}\label{take:represent}} ML is widely used to represent beliefs subsymbolically, especially via LLMs and NNs, but online learning and multi-agent knowledge enrichment are still lacking.}
\end{mybox}


\newcommand{\belieftabhead}{
	\textbf{\#} & \textbf{Method} & \textbf{Year} & \textbf{Strategy} & \textbf{ML} (\ref{item:rq4}) & \textbf{On. Learn.} (\ref{item:rq5}) & \textbf{Belief} (\ref{item:rq3}) & \textbf{Tech.} (\ref{item:rq6}) \\
	\hline\hline
}

\newcounter{BeliefMethod}
\setcounter{BeliefMethod}{1}
\newcommand{\newBeliefMethodIndex}{\theBeliefMethod\stepcounter{BeliefMethod}}

\begin{scriptsize}

\begin{longtable}{c|p{2.5cm}|c|c|c|c|c|c}
    \caption{
        Summary of \ac{ML}-enabled belief modules in \mlbdi{} agents. Legend: M=\emph{sub-symbolic belief modelling}, U=\emph{belief update}, E=\emph{knowledge enrichment}, V=\emph{vectorial}, B=\emph{bayesian}, EL=\emph{epistemic logic}, P=\emph{probabilistic}, UD=\emph{undefined}, L=\emph{language-based}, KG=\emph{knowledge graph}, CL=\emph{causal learning}.
    }
    \labeltab{belief-taxonomy}\\        
    \belieftabhead
    \endfirsthead
    \caption[]{Summary of \ac{ML}-enabled belief modules in \mlbdi{} agents (continued).}\\
    \belieftabhead
    \endhead
    \endlastfoot
    \newBeliefMethodIndex & \citesp{JianJim2007} & M & SONN & \cmark & V & \unavailable
    \\\hline
    \newBeliefMethodIndex & \citesp{LeeWsc2008} & U & BN & \xmark & B & \unavailable
    \\\hline
    \newBeliefMethodIndex & \citesp{LeeInforms2009} & U & RL & \cmark & B & \unavailable
    \\\hline
    \newBeliefMethodIndex & \citesp{LeeTomacs2010} & U & BN & \xmark & B & \unavailable
    \\\hline
    \newBeliefMethodIndex & \citesp{KimUoa2015} & U & BN & \xmark & B & \unavailable
    \\\hline
    \newBeliefMethodIndex & \citesp{ZhuangMatec2018} & U & NN & \xmark & V & \unavailable
    \\\hdashline
    \newBeliefMethodIndex & \citesp{RabinowitzIcml2018} & U & NN & \xmark & V & \unavailable
    \\\hline
    \newBeliefMethodIndex & \citesp{LeEmnlp2019} & M & LLM & \xmark & V & \tiny{\url{https://tinyurl.com/27osuant}} 
    \\\hdashline
    \newBeliefMethodIndex & \citesp{JaraCobs2019} & U & RL & \cmark & V & \unavailable
    \\\hline
    \newBeliefMethodIndex & \citesp{CuzzolinPsy2020} & U & RL & \cmark & V & \unavailable
    \\\hdashline
    \newBeliefMethodIndex & \citesp{DissingIjcai2020} & U & NN & \xmark & EL & \unavailable
    \\\hline
    \newBeliefMethodIndex & \citesp{MorenoArxiv2021} & E & RNN & \xmark & P & \unavailable
    \\\hline
    \newBeliefMethodIndex & \citesp{SclarIcml2022} & M & NN & \cmark & V & \tiny{\url{https://tinyurl.com/28ya5ycz}} 
    \\\hdashline
    \newBeliefMethodIndex & \citesp{HoTcs2022} & U & CL & \xmark & UD & \unavailable
    \\\hdashline
    \newBeliefMethodIndex & \citesp{HuangCorl2022} & U & LLM & \xmark & L & \unavailable
    \\\hline
    \newBeliefMethodIndex & \citesp{reflexion-neurips2023} & M & LLM & \xmark & L & \tiny{\url{https://tinyurl.com/2cdbhtmf}} 
    \\\hdashline
    \newBeliefMethodIndex & \citesp{ZhouAiide2023} & E & LLM & \cmark & KG & \unavailable
    \\\hdashline
    \newBeliefMethodIndex & \citesp{DipaloArxiv2023} & M & LLM & \cmark & V & \unavailable
    \\\hdashline
    \newBeliefMethodIndex & \citesp{PengAiide2023} & E & LLM & \xmark & KG & \unavailable
    \\\hdashline
    \newBeliefMethodIndex & \citesp{KumarHci2023} & U & NN & \xmark & V & \unavailable
    \\\hdashline
    \newBeliefMethodIndex & \citesp{SahaArxiv2023} & M & LLM & \xmark & V & \tiny{\url{https://tinyurl.com/2cr2842w}} 
    \\\hdashline
    \newBeliefMethodIndex & \citesp{ZhuArxiv2023} & U & LLM & \xmark & L & \tiny{\url{https://tinyurl.com/26fsgvbc}} 
    \\\hdashline
    \newBeliefMethodIndex & \citesp{react-iclr2023} & U & LLM & \xmark & L & \tiny{\url{https://tinyurl.com/24k4begz}} 
    \\\hdashline
    \newBeliefMethodIndex & \citesp{NascimentoAcsos2023} & M & LLM & \xmark & V & \unavailable
    \\\hdashline
    \newBeliefMethodIndex & \citesp{ZhaoPatterns2023} & U & NN & \xmark & V & \tiny{\url{https://tinyurl.com/24p93ldc}} 
    \\\hdashline
    \newBeliefMethodIndex & \citesp{generativeagents-2023} & U & LLM & \xmark & L & \tiny{\url{https://tinyurl.com/228m2eyp}} 
    \\\hdashline
    \newBeliefMethodIndex & \citesp{metaagents-2023} & U & LLM & \xmark & L & \unavailable
    \\\hdashline
    \newBeliefMethodIndex & \citesp{GaoSthree2023} & U & LLM & \xmark & L & \unavailable
    \\\hline
    \newBeliefMethodIndex & \citesp{LianWww2024} & M & LLM & \xmark & V & \tiny{\url{https://tinyurl.com/2d9ghoqe}} 
    \\\hdashline
    \newBeliefMethodIndex & \citesp{bdinl-aamas2024} & E & NN & \xmark & L & \tiny{\url{https://tinyurl.com/27uazxbb}} 
    \\\hdashline
    \newBeliefMethodIndex & \citesp{QianAcl2024} & E & LLM & \xmark & KG & \tiny{\url{https://tinyurl.com/ywzblfga}} 
    \\\hdashline
    \newBeliefMethodIndex & \citesp{WuAcl2024} & M & LLM & \xmark & L & \tiny{\url{https://tinyurl.com/26p2m7vq}} 
    \\\hdashline
    \newBeliefMethodIndex & \citesp{recmind-naacl2024} & U & LLM & \xmark & L & \unavailable
    \\\hdashline
    \newBeliefMethodIndex & \citesp{JinAcl2024} & M & LLM & \xmark & V & \tiny{\url{https://tinyurl.com/2a9mllem}} 
    \\\hdashline
    \newBeliefMethodIndex & \citesp{embodiedagentsllm-iclr2024} & U & LLM & \xmark & L & \tiny{\url{https://tinyurl.com/29kecyhe}} 
    \\\hdashline
    \newBeliefMethodIndex & \citesp{JiangWc2024} & U & LLM & \xmark & L & \tiny{\url{https://tinyurl.com/28hclv6o}} 
    \\\hdashline
    \newBeliefMethodIndex & \citesp{KannanIros2024} & M & LLM & \xmark & V & \tiny{\url{https://tinyurl.com/23tpb2d4}} 
    \\\hdashline
    \newBeliefMethodIndex & \citesp{NiEml2024} & M & LLM & \xmark & V & \tiny{\url{https://tinyurl.com/29384ch2}} 
    \\\hdashline
    \newBeliefMethodIndex & \citesp{retroformer-2024} & U & LLM & \xmark & L & \tiny{\url{https://tinyurl.com/247ysg6p}} 
    \\\hdashline
    \newBeliefMethodIndex & \citesp{metagpt-2024} & U & LLM & \xmark & L & \tiny{\url{https://tinyurl.com/29vyobuw}} 
    \\\hdashline
    \newBeliefMethodIndex & \citesp{Finrobot2024} & M & LLM & \xmark & V & \tiny{\url{https://tinyurl.com/292crmc3}} 
    \\\hline
    \newBeliefMethodIndex & \citesp{ZhaoArxiv2025} & M & LLM & \cmark & V & \unavailable
    \\\hdashline
    \newBeliefMethodIndex & \citesp{Recagent2025} & U & LLM & \xmark & L & \tiny{\url{https://tinyurl.com/2ygxwrcv}} 
    \\\hdashline
    \newBeliefMethodIndex & \citesp{Personalwab2025} & U & LLM & \xmark & L & \tiny{\url{https://tinyurl.com/22vyf3r8}} 
    \\\hdashline
    \newBeliefMethodIndex & \citesp{Wirelessagent2025} & U & LLM & \xmark & L & \tiny{\url{https://tinyurl.com/2bbt2ykd}} 
    \\\hdashline
    \newBeliefMethodIndex & \citesp{Codriving2025} & U & LLM & \xmark & L & \tiny{\url{https://tinyurl.com/2a5d64mr}} 
    \\\hline\hline
    
\end{longtable}

\end{scriptsize}

\begin{figure}
\centering
\begin{minipage}{.315\textwidth}
    \centering
    \includegraphics[width=\linewidth]{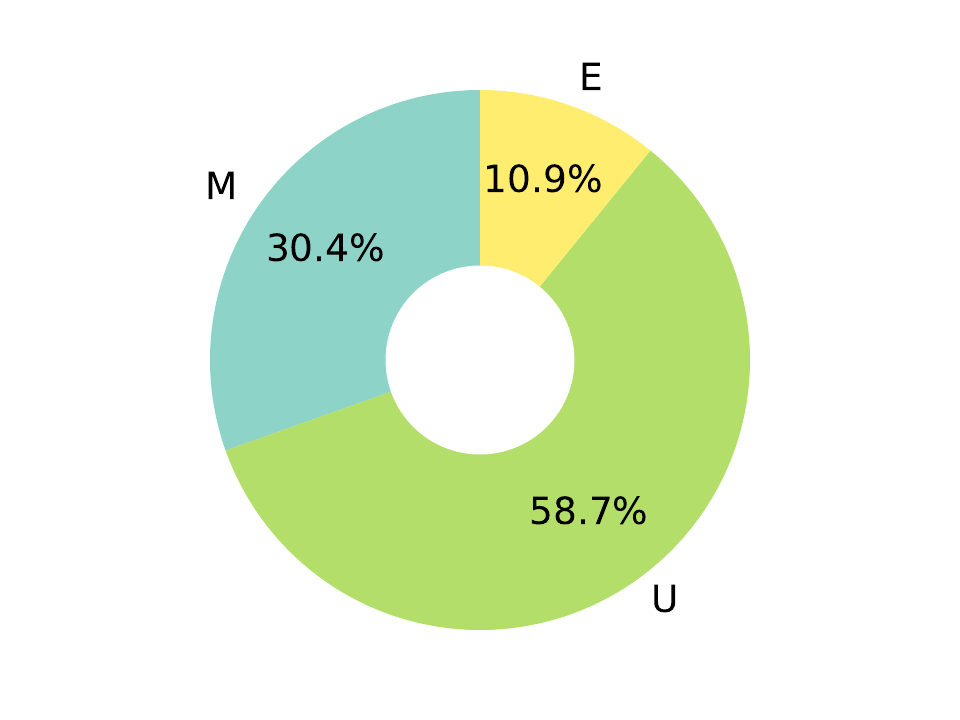}
    \captionof{figure}{Distribution of belief representation classes. See legend of \reftab{belief-taxonomy}.}
    \labelfig{belief_aim_plot}
\end{minipage}%
\hspace{0.2cm}
\begin{minipage}{.315\textwidth}
    \centering
    \includegraphics[width=\linewidth]{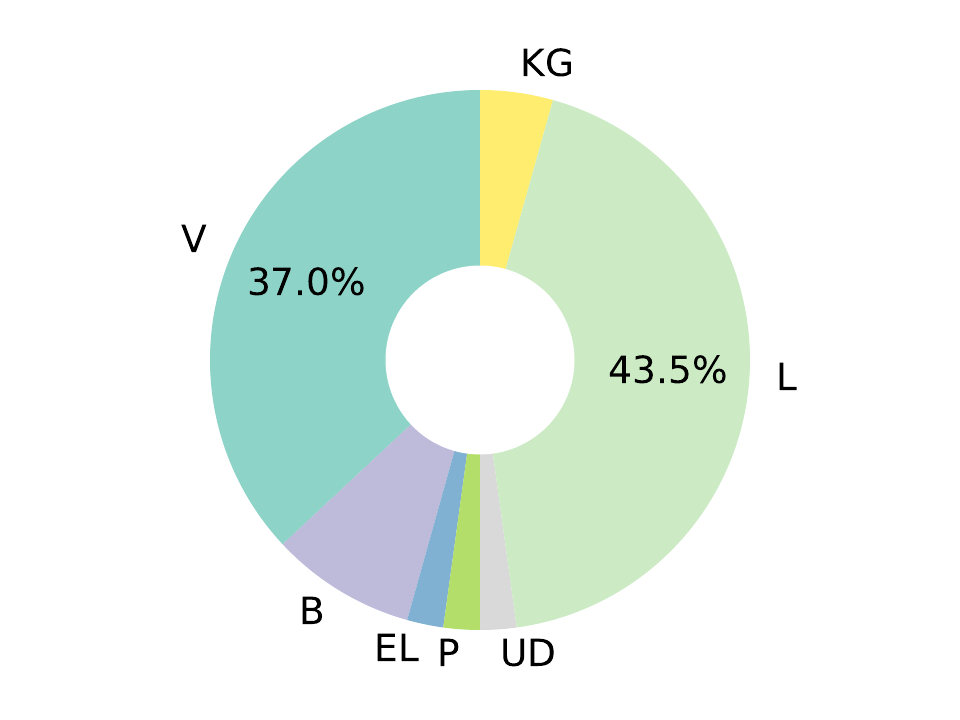}
    \captionof{figure}{Distribution of approaches to represent beliefs (\ref{item:rq3}). See legend of \reftab{belief-taxonomy}.}
    \labelfig{belief_struct_plot}
\end{minipage}
\hspace{0.2cm}
\begin{minipage}{.315\textwidth}
    \centering
    \includegraphics[width=\linewidth]{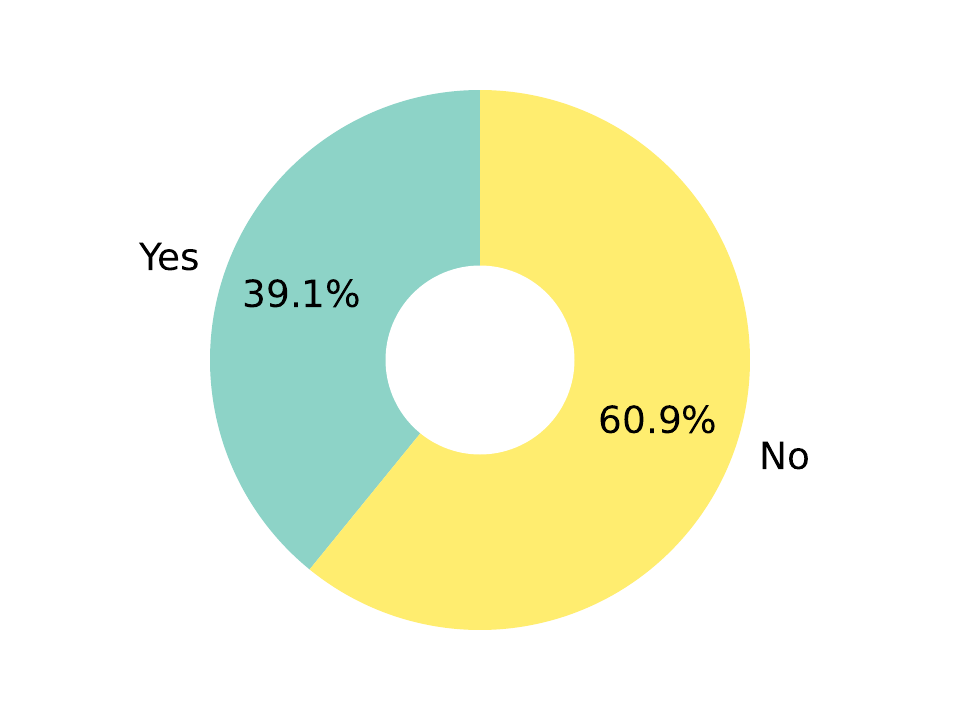}
    \captionof{figure}{Distribution of papers enabling online learning in belief representation (\ref{item:rq5}).}
    \labelfig{belief_learn_plot}
\end{minipage}
\end{figure}

\subsubsection*{\textbf{Opportunities and future directions}}
Recent research efforts focus heavily on integrating \ac{ML} approaches into rational agents to model beliefs in a \subsym{} fashion, enabling \ac{ML}-based belief processing.
While relevant, these approaches often overlook broader opportunities that arise from embedding \ac{ML} into the belief module.
\ac{ML}-based belief processing can extend beyond automatic extraction and update of beliefs from environmental feedback.
Moreover, knowledge enrichment proposals remain in early stages, typically relying on multi-party belief accumulation.
More extensive efforts are needed in \ac{ML}-based belief representation, especially in multi-agent settings where knowledge sharing is key to effective collaboration.

We envision future work focusing on \ac{ML}-driven belief enrichment by integrating techniques from knowledge graphs and ontologies.
Challenges like knowledge graph completion~\cite{ChenAccess2020,ShenKbs2022} and ontology construction~\cite{AswadiAir2020} are gaining traction.
Embedding these into \ac{BDI} agents \textendash{} where beliefs can be graph-structured \textendash{} could significantly extend the belief module.
Introducing \ac{LLM}-based reasoning into belief processing is a promising direction.
Some recent approaches leverage \ac{LLM} parametric knowledge to expand or construct structured knowledge via context-specific prompts~\cite{HaoAcl2023,llmoracles-kbs2025}.
However, key questions remain around reliability, leaving room for further investigation~\cite{PanTgdk2023}.

\Acf{ILP}~\cite{MuggletonAlt1990,KitzelmannAaip2009} also offers potential for belief enrichment in \mlbdi{} agents.
\ac{ILP} induces logic programs from data, generating new \sym{} hypotheses from positive, negative examples and background knowledge~\cite{CropperIjcai2020,CropperJair2022}.
In \ac{BDI} agents, background knowledge corresponds to the belief set, while examples can be drawn from interaction history, enabling belief completion via \ac{ILP}.
However, efficiently identifying examples for reliable knowledge induction remains an open challenge, as integrating \ac{ILP} into \mlbdi{} agents would require online learning and automatic evaluation of interaction outcomes\textemdash{}resembling an online learning \ac{ILP}-\ac{BDI} agent.

In \ac{ML}-based belief representation, minimal attention has been given to multi-agent contexts, where agents share environmental knowledge.
Yet, effective belief sharing is crucial for real-world agent deployment.
We foresee future research leveraging \ac{ML} to identify commonalities between agents' beliefs, thus allowing agents to incorporate others' experiences or knowledge and enhance collaboration.
We refer to this as the \emph{knowledge fusion} procedure, where a \mlbdi{} agent merges beliefs from others into a new valid belief set.
This mirrors knowledge fusion tasks in KG and ontology domains~\cite{NguyenInffus2020}, where recent works use \acp{LLM} to find commonalities across KGs on similar topics~\cite{YangArxiv2024}.
These approaches demonstrate the feasibility of knowledge fusion in \mlbdi{} agents.
However, inaccurate fusion may introduce unreliable or inconsistent beliefs, highlighting the need for consistency checking.
Therefore, we identify two key future research directions:
\begin{inlinelist}
    \item enabling \ac{ML}-based knowledge fusion in \ac{BDI} agents to support shared knowledge and collaboration, and
    \item developing efficient consistency checking techniques to detect conflicting beliefs across agents.
\end{inlinelist}

\section{Machine Learning and Desires}\labelsec{ml_desires}

This section presents our findings on the use of \ac{ML} models within the \ac{BDI} modules related to desire generation and representation.
Our analysis shows that this \ac{BDI} component remains vastly underexplored.
Only a few approaches consider learning and representing an agent's desires in a data-driven manner, and none of the selected papers address the option generation process for determining desires based on current intentions and beliefs.
We analyse in detail each paper presenting its workflow (thus answering~\ref{item:rq1}) and checking
\begin{inlinelist}
    \item the \ac{ML} model used by each approach (\ref{item:rq4});
    \item the proposed desire structure \textendash{} i.e., how desires are represented in the agent \textendash{} answering~\ref{item:rq3};
    \item if the proposed approach supports online learning of the \ac{ML} model inside the agent framework (\ref{item:rq5}); and
    \item if the proposed approach comes with a publicly available implementation, answering~\ref{item:rq6}.
\end{inlinelist} 

\subsection{Desire Representation}\labelssec{desire_rep}

In \ac{BDI} agents, desires represent the motivational state of the agent, defining objectives or situations the agent aims to accomplish.
Desire representation varies across agent frameworks, depending on application-specific requirements.
Using an \ac{ML}-enabled framework for desire representation presents a valuable opportunity for \mlbdi{} integration\textemdash{}enabling efficient \subsym{} processing, achievability checking, and more.
However, our literature analysis identifies only six works in this \mlbdi{} module, likely due to the complexity of modelling \sym{} knowledge through \subsym{} \ac{ML} approaches.
These approaches typically rely on \acp{NN} or other \subsym{} methods to represent desires implicitly via vectorial formats, most commonly \ac{NN} embeddings.
Many build on \ac{ToM} frameworks to model desires based on hypotheses about other agents' states.

\citeap{RabinowitzIcml2018} extend the \ac{ToM} model into a neural architecture, using three \acp{NN} to represent the agent's knowledge, state, and plans.
One model encodes the agent's motivational state \textendash{} its desires \textendash{} and adopted goals.
Meta-learning optimises each module, yielding a precise \subsym{} characterisation of desires.
\citeap{KumarHci2023} revisit this architecture, focusing on model development and applications.
In~\cite{JaraCobs2019}, the authors explore similarities between inverse \ac{RL} and \ac{ToM}, proposing to model desires by simulating a \ac{RL} model with hypothesised beliefs and desires.
Mental-state inference is achieved by inverting the model, effectively modelling the agent's desires.
Building on psychological findings about \emph{internal simulation} \textendash{} where humans predict others' actions by imagining their own \textendash{} \citeap{CuzzolinPsy2020} propose using simulations and \ac{RL} to build \ac{ToM} modules in robotic agents.
They suggest such simulations also support functions like episodic memory, counterfactual thinking, and future planning.
The synergy of \ac{RL} and simulation enables small neural components to represent desires, beliefs, and intentions.
In~\cite{OguntolaRoman2021}, an extended \ac{ToM} model replaces rule-based desire modules with a \ac{NN} trained on beliefs and actions to learn desires and intentions.

Frameworks like CoELA~\cite{embodiedagentsllm-iclr2024}, MetaAgents~\cite{metaagents-2023}, and others~\cite{generativeagents-2023,metagpt-2024,GaoSthree2023} use memory modules to store agent knowledge and experience\textemdash{}including other agents in a \ac{MAS}.
This knowledge informs planning and action generation, implicitly modelling desires to avoid conflicts.
However, most works lack detailed discussion on desire modelling, warranting further investigation.
A similar pattern appears in MechAgents~\cite{NiEml2024}, Smart-LLM~\cite{KannanIros2024}, and FinRobot~\cite{Finrobot2024}, where a \ac{LLM}-based coordinator agent encodes desires and intentions to optimise plans.
These agents lack memory modules, so desires and intentions are represented vectorially in the \ac{LLM} embedding space.

Recently, \citeap{FreringEaai2025} and~\citeap{chatbdi-2025} propose frameworks where \ac{BDI} agents interact with humans via \ac{LLM}-based text processing to receive attainable desires \textendash{} i.e., goals \textemdash{} and explain actions through \emph{chattification}.
Desires are represented in textual or \sym{} form and used to identify plans via purely \sym{} processing, without plan optimisation or option generation.

\reftab{desire-taxonomy} summarises the main features of each analysed work, while \reffigfromto{desire_struct_plot}{desire_learn_plot} overview the paper distribution.

\begin{mybox}
\phantomsection
{{\bf \faSearch\ \textit{Quick Takeaway:}\label{take:desire}} Desire representation via ML is limited, mostly relying on subsymbolic embeddings or ToM frameworks, with minimal support for symbolic reasoning.}
\end{mybox}


\newcommand{\desiretabhead}{
	\textbf{\#} & \textbf{Method} & \textbf{Year} & \textbf{ML} (\ref{item:rq4}) & \textbf{On. Learn.} (\ref{item:rq5}) & \textbf{Desire} (\ref{item:rq3}) & \textbf{Tech.} (\ref{item:rq6}) \\
	\hline\hline
}

\newcounter{DesireMethod}
\setcounter{DesireMethod}{1}
\newcommand{\newDesireMethodIndex}{\theDesireMethod\stepcounter{DesireMethod}}

\begin{scriptsize}

\begin{longtable}{c|p{3cm}|c|c|c|c|c}
    \caption{
        Summary of \ac{ML}-enabled desire representation in \mlbdi{} agents. Legend: V=\emph{vectorial}, L=\emph{language}, J=\emph{jason}.
    }
    \labeltab{desire-taxonomy}\\        
    \desiretabhead
    \endfirsthead
    \caption[]{Summary of \ac{ML}-enabled desire representation in \mlbdi{} agents (continued).}\\
    \desiretabhead
    \endhead
    \endlastfoot
    \newDesireMethodIndex & \citesp{RabinowitzIcml2018} & NN & \xmark & V & \unavailable
    \\\hline
    \newDesireMethodIndex & \citesp{JaraCobs2019} & RL & \cmark & V & \unavailable
    \\\hline
    \newDesireMethodIndex & \citesp{CuzzolinPsy2020} & RL & \cmark & V & \unavailable
    \\\hline
    \newDesireMethodIndex & \citesp{OguntolaRoman2021} & NN & \xmark & V & \unavailable
    \\\hline
    \newDesireMethodIndex & \citesp{KumarHci2023} & NN & \xmark & V & \unavailable
    \\\hdashline
    \newDesireMethodIndex & \citesp{generativeagents-2023} & LLM & \xmark & L & \tiny{\url{https://tinyurl.com/228m2eyp}} 
    \\\hdashline
    \newDesireMethodIndex & \citesp{metaagents-2023} & LLM & \xmark & L & \unavailable
    \\\hdashline
    \newDesireMethodIndex & \citesp{GaoSthree2023} & LLM & \xmark & L & \unavailable
    \\\hline
    \newDesireMethodIndex & \citesp{embodiedagentsllm-iclr2024} & LLM & \xmark & L & \tiny{\url{https://tinyurl.com/29kecyhe}} 
    \\\hdashline
    \newDesireMethodIndex & \citesp{KannanIros2024} & LLM & \xmark & V & \tiny{\url{https://tinyurl.com/23tpb2d4}} 
    \\\hdashline
    \newDesireMethodIndex & \citesp{NiEml2024} & LLM & \xmark & V & \tiny{\url{https://tinyurl.com/29384ch2}} 
    \\\hdashline
    \newDesireMethodIndex & \citesp{metagpt-2024} & LLM & \xmark & V & \tiny{\url{https://tinyurl.com/29vyobuw}} 
    \\\hdashline
    \newDesireMethodIndex & \citesp{Finrobot2024} & LLM & \xmark & V & \tiny{\url{https://tinyurl.com/292crmc3}} 
    \\\hline
    \newDesireMethodIndex & \citesp{FreringEaai2025} & LLM & \xmark & L & \unavailable
    \\\hdashline
    \newDesireMethodIndex & \citesp{chatbdi-2025} & LLM & \xmark & J & \tiny{\url{https://tinyurl.com/263gjod6}} 
    \\\hline\hline

\end{longtable}

\end{scriptsize}

\begin{figure}[!t]
\centering
\begin{minipage}{.315\textwidth}
    \centering
    \includegraphics[width=\linewidth]{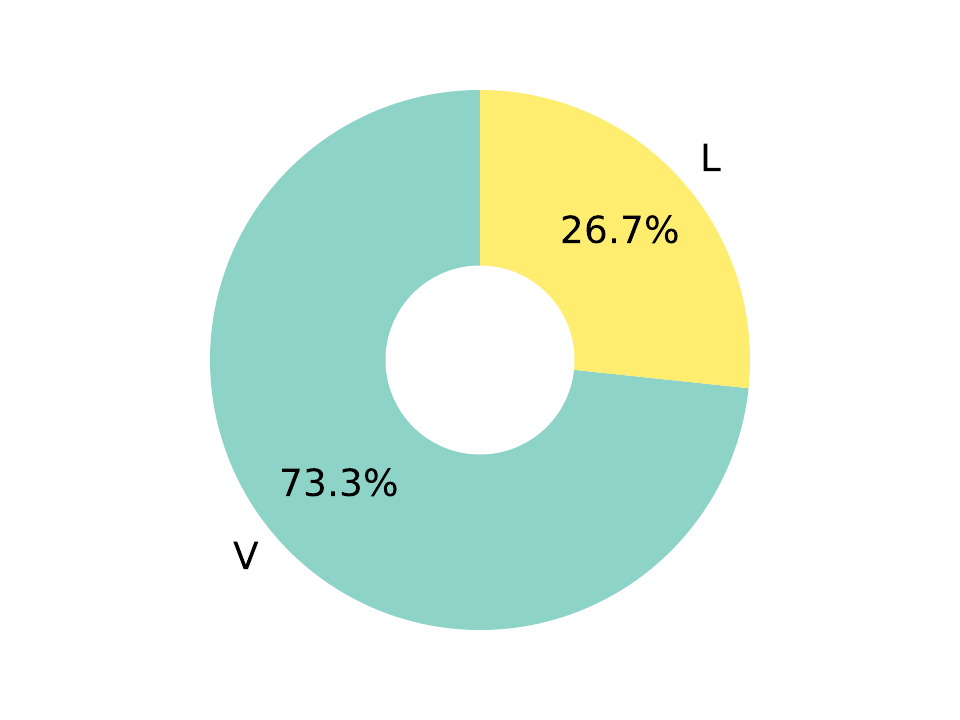}
    \captionof{figure}{Distribution of approaches to represent desires (\ref{item:rq3}). See legend of \reftab{desire-taxonomy}.}
    \labelfig{desire_struct_plot}
\end{minipage}
\hspace{1.5cm}
\begin{minipage}{.315\textwidth}
    \centering
    \includegraphics[width=\linewidth]{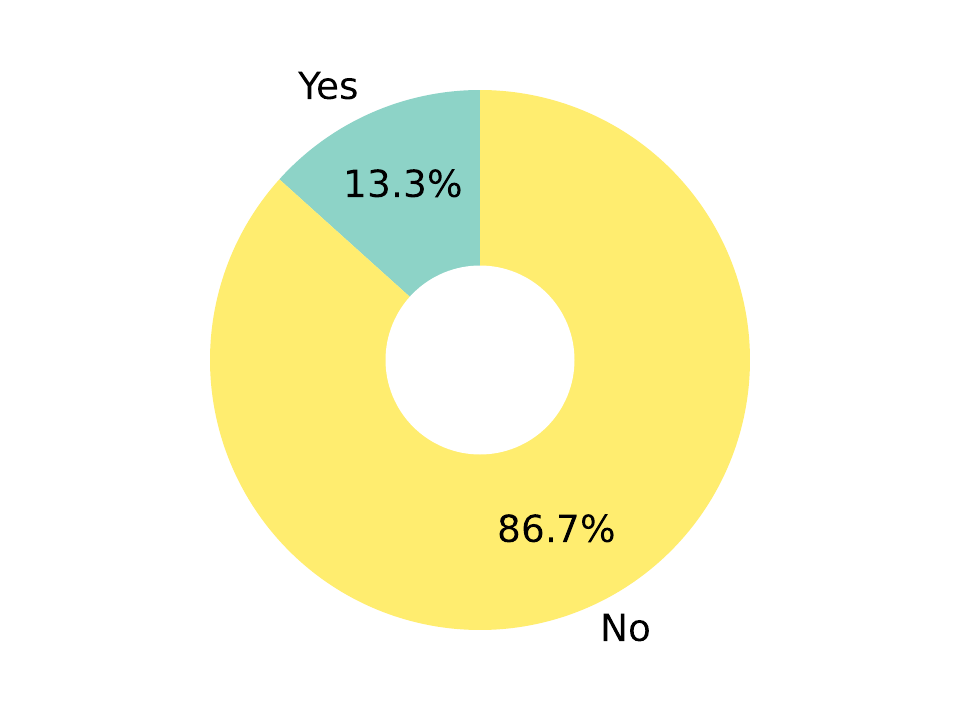}
    \captionof{figure}{Distribution of papers enabling online learning in desires representation (\ref{item:rq5}).}
    \labelfig{desire_learn_plot}
\end{minipage}
\end{figure}

\subsubsection*{\textbf{Opportunities and future directions}}
Given the limited number and recency of papers in this category, we anticipate increased research efforts in the future.
Using \subsym{} approaches to map and process agents' desires presents a compelling challenge, increasingly gaining traction\textemdash{}especially in light of works on \subsym{} belief representation (see \refssec{belief_rep}).
We foresee more advanced \ac{ML}-based methods where desires are not only represented via \subsym{} embeddings but also manipulated through vectorial operations.
Such representations could support tasks like \emph{conflict resolution}, \emph{management}, and \emph{achievability checking}.

In conflict resolution, given a set of desires, the agent framework may use \ac{ML} models to detect inconsistencies\textemdash{}either within its own desires or across those of other agents.
\ac{ML} could be used to:
\begin{inlinelist}
    \item identify and resolve conflicting desires end-to-end, or
    \item accelerate conflict detection prior to symbolic resolution.
\end{inlinelist}
Embedding-based representations allow for measuring distance and overlap between desires, enabling fast geometric conflict detection without relying on time-consuming \sym{} execution.

For \emph{achievability checking}, we propose using \ac{ML} to assess whether new desires are feasible within the agent's environment.
Not all generated desires are achievable, especially in dynamic or complex settings.
Thus, automated achievability checks are crucial for deploying \ac{BDI} agents in real-world scenarios.
Traditional checking is often cumbersome, requiring predictions of environmental evolution and \ac{ML} can support preemptive filtering of unfeasible desires by predicting environmental dynamics based on the agent's experience.

\subsection{Option Generation}\labelssec{desire_gen}

Whenever a \ac{BDI} agent deliberates, it begins by identifying its available options, generating a set of possible alternatives.
The option generation function defines how the agent determines viable intentions based on its current internal state.
Given the purely \sym{} nature of this task and the complexity of generating \sym{} and semantically valid concepts via \ac{ML}, it is unsurprising that our literature analysis finds no relevant papers in this area\textemdash{}see \reffig{paper_distribution}.

The process should avoid generating intentions that conflict with the agent's beliefs or goals.
Options not targeting achievable (sub)goals should be excluded, and those clashing with the agent's beliefs about the environment are invalid, as they correspond to unfeasible actions.
Integrating \ac{ML} into option generation is challenging due to the probabilistic nature of \ac{ML} models.
For instance, hallucinations from \acp{LLM} may lead to the agent considering irrelevant intentions.
Nonetheless, with the growing popularity of \subsym{} representations of beliefs, desires, and intentions~\cite{JinAcl2024,embodiedagentsllm-iclr2024,NottinghamIcml2023}, using \ac{ML} for option generation may become increasingly important.
Many agent frameworks now rely on purely \subsym{} processes where option generation is implicit.
Thus, employing \ac{ML} to generate options could allow this key reasoning step to be explicitly represented, rather than embedded within an end-to-end \ac{ML} pipeline.

\begin{mybox}
\phantomsection
{{\bf \faSearch\ \textit{Quick Takeaway:}\label{take:options}} No surveyed works explicitly address ML-based option generation, possibly due to the unreliability of ML models in the BDI deliberation context.}
\end{mybox}

\subsubsection*{\textbf{Opportunities and future directions}}

The option generation process of a \ac{BDI} agent is a delicate part of its reasoning cycle, typically requiring high reliability.
We foresee a promising trend in applying generative \ac{ML} approaches \textendash{} such as \ac{LLM} \textendash{} to generate partially consistent or probabilistic intentions within probabilistic and para-consistent agent frameworks~\cite{DuninKepliczAmai2024}.
Using generative \ac{ML} to automatically produce potential intentions may be particularly useful in frameworks relying on non-\sym{} representations of beliefs and intentions.
For instance, in conversational agents where beliefs, desires, and intentions are expressed in textual form, leveraging \ac{LLM} for option generation could be effective.

We also highlight the potential of combining a \ac{ML}-based generation process with a \sym{}-based validity check as a valuable research direction.
Such coupling could ensure that generated options are both expressive and verifiable.
Hence, future frameworks may integrate \ac{ML} model(s) with \sym{} verification techniques to define \neusym{} option generation modules within \mlbdi{} agents.

\section{Machine Learning and Intentions}\labelsec{ml_intentions}

In this section, we present our literature findings on the integration of \ac{ML} approaches into \ac{BDI} modules related to agent intentions.
In a \ac{BDI} agent, intentions represent its deliberative state \textendash{} what the agent has chosen to do \textendash{} and correspond to desires the agent has committed to, along with instantiated plans~\cite{bdiplanning-ker30}.
Intentions require a filtering or selection process to choose which desires to commit to, and a plan generation or selection process to implement actions.
We therefore focus on intention representation, generation, and planning.
Our analysis shows that \ac{ML} approaches for intention representation and planning are gaining popularity, likely due to their ability to learn the agent's preferred acting process automatically.
By contrast, intention filtering remains underexplored and warrants greater attention.
We analyse in detail each paper presenting its workflow (thus answering~\ref{item:rq1}) and checking
\begin{inlinelist}
    \item the \ac{ML} model used by each approach (\ref{item:rq4});
    \item the intentions and/or plan structure \textendash{} i.e., how intentions and plans are represented in the agent \textendash{} answering~\ref{item:rq3};
    \item if the proposed approach supports online learning of the \ac{ML} model inside the agent framework (\ref{item:rq5}); and
    \item if the proposed approach comes with a publicly available implementation, answering~\ref{item:rq6}.
\end{inlinelist} 

\subsection{Intentions Representation}\labelssec{intention_rep}

Our analysis reveals a strong connection between existing approaches to \ac{ML}-based intention and desire representation\textemdash{}see \refssec{desire_rep}.
Several works use \acp{NN} or other \subsym{} mechanisms to implicitly model intentions via vectorial representations, typically \ac{NN} embeddings.
Some also build on \ac{ToM} frameworks to model intentions based on hypotheses about other agents' states.
A few papers target both desire and intention representation~\cite{RabinowitzIcml2018,OguntolaRoman2021,embodiedagentsllm-iclr2024,GaoSthree2023}, using either separate \ac{ML} modules or a shared \ac{NN}.

\citeap{TahboubJirs2006} focuses on human-robot interaction, using \ac{BN} to recognise human intentions, which are then used to model the robot's own intentions and guide planning.
Similarly,~\cite{DiaconescuPloscb2014} mimic human intention recognition using hierarchical Bayesian networks.
\citeap{NguyenRoman2018} tackle self-intention generation via an action learning mechanism using deep Q-networks~\cite{MnihNature2015} and R\ac{NN}, where the latter encodes context embeddings for intention prediction.
Some approaches use \acp{LLM} to implicitly construct intentions in coordinator agents of \ac{MAS}~\cite{embodiedagentsllm-iclr2024,KannanIros2024,NiEml2024}.
Although not explicitly linked to individual agents' intentions, the \ac{LLM}-based plan generation process accounts for them to avoid conflicts.
Intentions may be represented in plain text~\cite{embodiedagentsllm-iclr2024,generativeagents-2023,metaagents-2023} or embedded within the \ac{LLM} representation~\cite{KannanIros2024,NiEml2024,metagpt-2024}.
The CoDrivingLLM framework~\cite{Codriving2025} shares intentions among agents and uses an \ac{LLM} engine for conflict negotiation and resolution.

In \ac{MAS} settings, \ac{ToM}-based intention representation extends the \ac{ToM} model into neural form for integration into agents.
Several works model an agent's intentions using \acp{NN}, based on its \ac{ToM} about others.
Neural \ac{ToM} models may include modules to represent other agents' intentions \subsym{}ally, enabling dual-layer intention modelling.
\citeap{RabinowitzIcml2018} first propose such a model, using a \ac{NN} sub-module to represent the agent's intentions and plans.
This is extended in~\cite{OguntolaRoman2021} and~\cite{KumarHci2023}, where intention learning depends on the agent's beliefs and actions.
\citeap{JaraCobs2019} suggests modelling intentions by inverting a simulated \ac{RL} model based on hypothesised beliefs and desires.
Similarly,~\cite{CuzzolinPsy2020} build on the human internal simulation mechanism, proposing that combining \ac{RL} and simulation enables neural components in agents to understand others' beliefs, desires, and intentions.

\reftab{intention-taxonomy} summarises the main features of each analysed work, while \reffigfromto{intention_struct_plot}{intention_learn_plot} provides an overview of the paper distribution.

\begin{mybox}
\phantomsection
{{\bf \faSearch\ \textit{Quick Takeaway:}\label{take:intend}} ML-based intention representation is gaining traction, often coupled with desire representation, but remains mostly implicit and lacks transparency.}
\end{mybox}


\newcommand{\intentiontabhead}{
	\textbf{\#} & \textbf{Method} & \textbf{Year} & \textbf{ML} (\ref{item:rq4}) & \textbf{On. Learn.} (\ref{item:rq5}) & \textbf{Intention} (\ref{item:rq3}) & \textbf{Tech.} (\ref{item:rq6}) \\
	\hline\hline
}

\newcounter{IntentionMethod}
\setcounter{IntentionMethod}{1}
\newcommand{\newIntentionMethodIndex}{\theIntentionMethod\stepcounter{IntentionMethod}}

\begin{scriptsize}

\begin{longtable}{c|p{3cm}|c|c|c|c|c}
    \caption{
        Summary of \ac{ML}-enabled intention representation in \mlbdi{} agents. Legend: P=\emph{probabilistic}, V=\emph{vectorial}, L=\emph{language-based}.
    }
    \labeltab{intention-taxonomy}\\        
    \intentiontabhead
    \endfirsthead
    \caption[]{Summary of \ac{ML}-enabled intention representation in \mlbdi{} agents (continued).}\\
    \intentiontabhead
    \endhead
    \endlastfoot
    \newIntentionMethodIndex & \citesp{TahboubJirs2006} & BN & \xmark & P & \unavailable
    \\\hline
    \newIntentionMethodIndex & \citesp{DiaconescuPloscb2014} & BN & \xmark & P & \unavailable
    \\\hline
    \newIntentionMethodIndex & \citesp{NguyenRoman2018} & RL & \cmark & V & \unavailable
    \\\hdashline
    \newIntentionMethodIndex & \citesp{RabinowitzIcml2018} & NN & \xmark & V & \unavailable
    \\\hline
    \newDesireMethodIndex & \citesp{JaraCobs2019} & RL & \cmark & V & \unavailable
    \\\hline
    \newDesireMethodIndex & \citesp{CuzzolinPsy2020} & RL & \cmark & V & \unavailable
    \\\hline
    \newDesireMethodIndex & \citesp{OguntolaRoman2021} & NN & \xmark & V & \unavailable
    \\\hline
    \newDesireMethodIndex & \citesp{KumarHci2023} & NN & \xmark & V & \unavailable
    \\\hdashline
    \newIntentionMethodIndex & \citesp{generativeagents-2023} & LLM & \xmark & L & \tiny{\url{https://tinyurl.com/228m2eyp}} 
    \\\hdashline
    \newIntentionMethodIndex & \citesp{metaagents-2023} & LLM & \xmark & L & \unavailable
    \\\hdashline
    \newIntentionMethodIndex & \citesp{GaoSthree2023} & LLM & \xmark & L & \unavailable
    \\\hline
    \newIntentionMethodIndex & \citesp{embodiedagentsllm-iclr2024} & LLM & \xmark & L & \tiny{\url{https://tinyurl.com/29kecyhe}} 
    \\\hdashline
    \newIntentionMethodIndex & \citesp{KannanIros2024} & LLM & \xmark & V & \tiny{\url{https://tinyurl.com/23tpb2d4}} 
    \\\hdashline
    \newIntentionMethodIndex & \citesp{NiEml2024} & LLM & \xmark & V & \tiny{\url{https://tinyurl.com/29384ch2}} 
    \\\hdashline
    \newIntentionMethodIndex & \citesp{metagpt-2024} & LLM & \xmark & V & \tiny{\url{https://tinyurl.com/29vyobuw}} 
    \\\hdashline
    \newIntentionMethodIndex & \citesp{Finrobot2024} & LLM & \xmark & V & \tiny{\url{https://tinyurl.com/292crmc3}} 
    \\\hline
    \newIntentionMethodIndex & \citesp{Codriving2025} & LLM & \xmark & L & \tiny{\url{https://tinyurl.com/2a5d64mr}} 
    \\\hline\hline

\end{longtable}

\end{scriptsize}

\begin{figure}[!b]
\centering
\begin{minipage}{.315\textwidth}
    \centering
    \includegraphics[width=\linewidth]{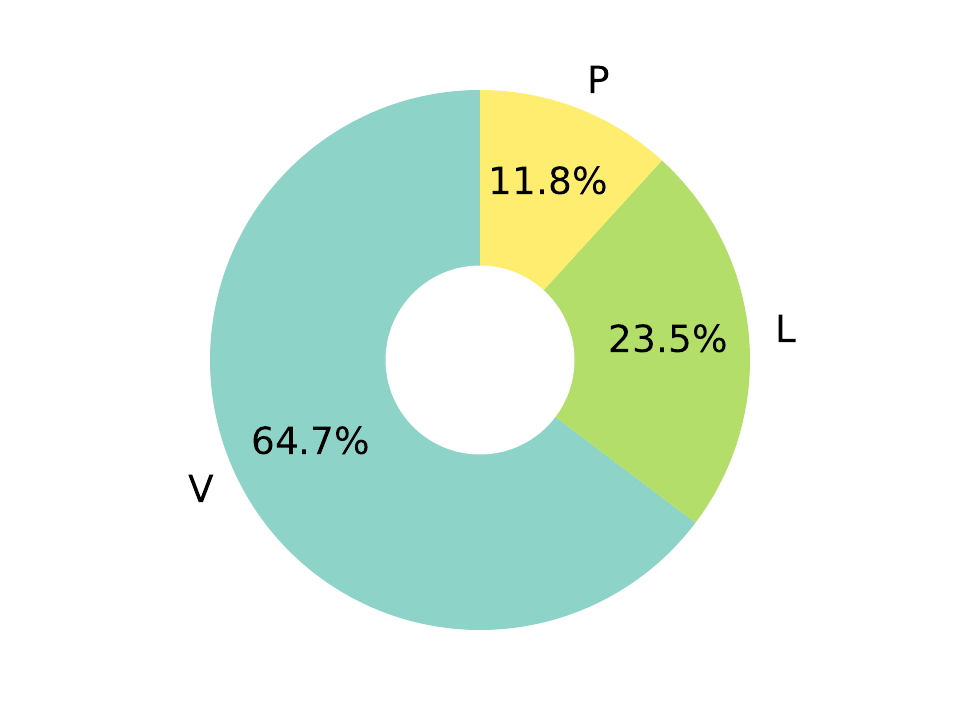}
    \captionof{figure}{Distribution of approaches to represent intentions (\ref{item:rq3})\textemdash{}see legend of \reftab{intention-taxonomy}.}
    \labelfig{intention_struct_plot}
\end{minipage}
\hspace{1.5cm}
\begin{minipage}{.315\textwidth}
    \centering
    \includegraphics[width=\linewidth]{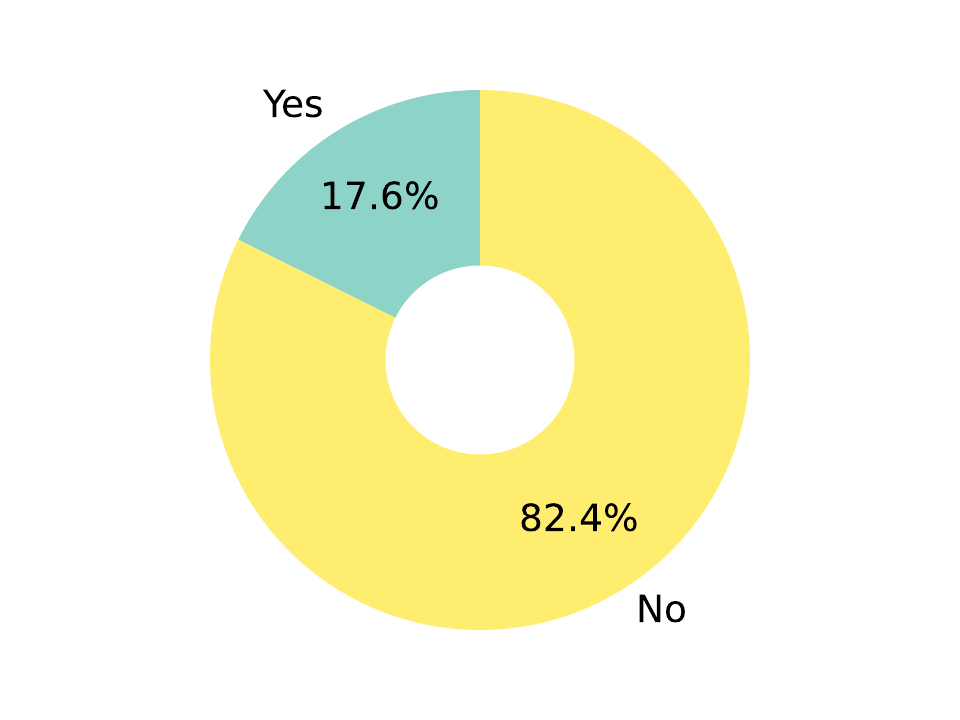}
    \captionof{figure}{Distribution of papers enabling online learning in intentions representation (\ref{item:rq5}).}
    \labelfig{intention_learn_plot}
\end{minipage}
\end{figure}

\subsubsection*{\textbf{Opportunities and future directions}}

With the rapid rise of \subsym{}-based agent frameworks, we expect future research to increasingly leverage \ac{ML} models to represent agent intentions \textendash{} either implicitly or explicitly \textendash{} within \ac{BDI} or \ac{BDI}-like frameworks.
The growing number of papers coupling desire and intention representation via \ac{NN} embeddings supports this trend and reflects the community's interest.
However, most current approaches are preliminary and rely on implicit intention representation.
While efficient, implicit modelling raises transparency and reliability concerns, as it prevents validation of the agent's intentions.
We therefore anticipate future work to focus on explicit intention representation, potentially using generative \ac{AI} models to construct verifiable \sym{} intentions.

Using \ac{ML} to model intentions may also support other complex tasks central to intelligent agent behaviour.
For instance, \ac{ML} could assist in checking whether intentions comply with environmental laws or norms.
This \emph{norm-checking} process is typically demanding and requires full environmental knowledge.
\ac{ML} techniques could help predict environmental evolution and streamline this process.
Similarly, \ac{ML} may identify when, based on context, an intention should be temporarily suspended.
This \emph{smart start and stop} mechanism depends on modelling the environment and agent state, requiring data-driven \ac{ML} integration.

Finally, we highlight the potential of \ac{ML} for \emph{intention sorting}, ranking intentions by relevance or expected reward.
Here, \ac{ML} enables rapid estimation of intention rewards, supporting efficient prioritisation.

\subsection{Intention Filtering}\labelssec{intention_filt}

The intention filtering process is a key component of a \ac{BDI} agent, enabling the refinement of selected options into current intentions, while considering beliefs and prior intentions.
Using a filter function, unrealistic or currently unachievable intentions \textendash{} based on the agent or environment status \textendash{} are excluded.
This ensures realistic and achievable intentions and supports coordination between agents\textemdash{}e.g., avoiding intentions that conflict with other agents' states or goals.
Despite its importance, our literature analysis identifies only three papers using \ac{ML} within this module.
This gap may stem from the difficulty of designing data-driven filtering functions tailored to dynamic agent and environment conditions.
The available frameworks are recent and rely on complex \acp{NN} to perform implicit filtering over \subsym{} representations of intentions, desires, and beliefs.

The DECKARD agent framework~\cite{NottinghamIcml2023} uses an \ac{LLM} to generate plans based on goals, which are decomposed into sub-goals by the same model.
This decomposition implicitly performs intention filtering, forming an abstract world model that includes intentions, later verified and refined via \ac{RL}.
\citeap{IchidaSac2023} introduce \ac{NN}-based language models in a conversational \ac{BDI} agent to interpret human input and update beliefs.
Intentions are selected to generate optimal suggestions for the user.
This framework is extended in~\cite{bdinl-aamas2024}, proposing a natural language-based \ac{BDI} agent where beliefs, plans, and intentions are expressed in text.
Intentions are processed and filtered using \ac{NN} modules to identify optimal actions in a \subsym{} manner.

\reftab{filter-taxonomy} summarises the main features of the analysed works.

\begin{mybox}
\phantomsection
{{\bf \faSearch\ \textit{Quick Takeaway:}\label{take:filter}} ML-driven intention filtering is rare and typically embedded in end-to-end pipelines, with limited support for symbolic validation or multi-agent coordination.}
\end{mybox}


\newcommand{\filtertabhead}{
	\textbf{\#} & \textbf{Method} & \textbf{Year} & \textbf{ML} (\ref{item:rq4}) & \textbf{On. Learn.} (\ref{item:rq5}) & \textbf{Intention} (\ref{item:rq3}) & \textbf{Tech.} (\ref{item:rq6}) \\
	\hline\hline
}

\newcounter{FilterMethod}
\setcounter{FilterMethod}{1}
\newcommand{\newFilterMethodIndex}{\theFilterMethod\stepcounter{FilterMethod}}

\begin{scriptsize}

\begin{longtable}{c|p{3cm}|c|c|c|c|c}
    \caption{
        Summary of \ac{ML}-enabled intention filtering in \mlbdi{} agents. Legend: V=\emph{vectorial}, PAS=\emph{python agent-speak}, L=\emph{language-based}.
    }
    \labeltab{filter-taxonomy}\\        
    \filtertabhead
    \endfirsthead
    \caption[]{Summary of \ac{ML}-enabled intention filtering in \mlbdi{} agents (continued).}\\
    \filtertabhead
    \endhead
    \endlastfoot
    \newFilterMethodIndex & \citesp{NottinghamIcml2023} & RL & \cmark & V & \tiny{\url{https://tinyurl.com/28555kln}} 
    \\\hdashline
    \newBrfMethodIndex & \citesp{IchidaSac2023} & NN & \xmark & PAS & \unavailable
    \\\hline
    \newBrfMethodIndex & \citesp{bdinl-aamas2024} & NN & \xmark & L & \tiny{\url{https://tinyurl.com/27uazxbb}} 
    \\\hline\hline

\end{longtable}

\end{scriptsize}

\subsubsection*{\textbf{Opportunities and future directions}}
%
Given the recency of approaches focusing on \ac{ML}-enabled intention filtering in \ac{BDI} or \ac{BDI}-like agents, we believe future research will increasingly explore this area.
While current methods rely on \subsym{} models to perform filtering in an end-to-end fashion, we envision future proposals using \ac{ML} to accelerate reliable \sym{} intention filtering procedures.
Purely \ac{ML}-based filtering may compromise reliability, being prone to hallucinations~\cite{HuangArxiv2023Hallucinations} and out-of-distribution issues~\cite{GawlikowskiAir2023}.
We also foresee \ac{ML} enabling deliberation in intention filtering and selection within \ac{MAS}.
\ac{ML} models could help agents reach agreement \textendash{} either directly or via their \ac{ToM} models \textendash{} on mental states and intentions.
For instance, when agents communicate their desires and intentions, each could filter its own intentions to avoid conflicts with others' actions or goals.
In this context, \ac{ML}-based argumentation~\cite{WangAcl2018} or argument mining~\cite{LawrenceColing2019} could support smart deliberation.
We therefore anticipate the integration of \ac{ML}-driven argumentation into \sym{} intention filtering pipelines.

\subsection{Planning}\labelssec{planning}

In \ac{BDI} agents, plans are essential for achieving goals, as they define sequences of actions \textendash{} usually represented as recipes \textendash{} that the agent performs to fulfil its intentions.
Plans may include sub-plans and are often only partially defined, with details refined as the agent interacts with the environment.
Given the central role of planning in intelligent agents, our survey finds that \ac{ML}-based planning solutions are by far the most common approach for integrating \ac{ML} into \ac{BDI} agents.
We identify \plancount{} works falling within this module.
These works can be broadly categorised into three classes of approaches, namely:
\begin{description}
    \item[Plan generation] in which the agent leverages \ac{ML} to construct from scratch one or more plan(s) of actions \textendash{} possibly composed of sub-plans \textendash{} given its beliefs, desires, and intentions.
    \item[Plan selection] in which \ac{ML} models are used to identify the best plan \textendash{} depending on the agent's intentions \textendash{} from a pool of available plans.
    \item[Plan optimisation] in which the agent uses \ac{ML} models to modify one or more already available plans to fit the evolution of the environment state or the agent's own updates.
\end{description}
Accordingly, we delve into the details of each category, while \reftab{plan-taxonomy} summarises our findings for each paper and~\reffigfromto{plan_aim_plot}{plan_learn_plot} provide an overview of the papers distribution.

\subsubsection{Plan generation}
With the growing popularity of agent frameworks using \acp{LLM}~\cite{agentsllm-fcs18}, \ac{ML}-based plan generation is becoming increasingly prominent.
Of the \plancount{} works, \plangencount{} focus on plan generation (see \reftab{plan-taxonomy}).
Many rely on \acp{LLM} to model planning and generate textual action plans, differing in how the reasoning engine is defined and used.
For instance, in~\cite{chainofthoughts-neurips2022}, Chain of Thought (CoT) is used by prompting reasoning steps into the \ac{LLM} to generate agent's plans, showcasing planning improvements over prompting the model using naive instructions such as done in~\cite{DipaloArxiv2023}.
Similarly, HuggingGPT~\cite{hugginggpt-neurips2023} uses ChatGPT\footnote{\url{https://openai.com/index/chatgpt/}} for task planning.
Tree-of-Thought (ToT)~\cite{treeofthoughts-neurips2023} powers RecMind~\cite{recmind-naacl2024}, where the \ac{LLM} handles both planning and belief processing.
Meanwhile, some works use \acp{LLM} as zero-shot planners~\cite{HuangIcml2022}, generating multiple next steps and selecting the best via distance metrics.
For example, \citeap{GramopadhyeIros2023} combine CoT and zero-shot planning with example-based prompting, while \citeap{HaoEmnlp2023} add a world model to simulate plan outcomes.
However, only a few of these works generate plans in rigorous \sym{} formats~\cite{SchulzIjpeds2025,CiattoEcai2025}.

Advanced frameworks refine \ac{LLM}-generated plans using feedback loops, for instance:
\begin{inlinelist}
    \item \citeap{HuangCorl2022} propose grounded closed-loop planning with perception models and language-conditioned robot skills.
    \item SayPlan~\cite{RanaCorl2023} iteratively refines plans using scene graph feedback.
    \item \citeap{SongIccv2023} update plans dynamically when actions fail.
    \item PET~\cite{WuArxiv2023} combines planning with modules for object masking and sub-task tracking.
    \item SwiftSage~\cite{swiftsage-neurips2023} emulate deliberation and plan goals.
    \item DECKARD~\cite{NottinghamIcml2023} generates plans via \ac{LLM}, verifies them with \ac{RL}, and updates them accordingly.
\end{inlinelist}

The textual representation of an agent's plans can also be rendered arbitrarily complex to enable more effective planning.
For example, in Voyager~\cite{voyager-tmlr2024}, the agent stores skills in a textual memory used as a plan library.
Meanwhile, frameworks like CoELA~\cite{embodiedagentsllm-iclr2024}, MetaAgents~\cite{metaagents-2023}, and others~\cite{generativeagents-2023,metagpt-2024} use memory modules to store world knowledge and other agents' states, informing planning via \ac{LLM}.
Finally, application context also influences how agents use \acp{LLM} for planning, as the applications span over recommendation agents~\cite{HuangArxiv2023,LianWww2024}, industrial settings~\cite{XiaEtfa2023}, electronic design~\cite{chateda-tcad43}, human-computer interaction~\cite{RuanArxiv2023,tptuv2-2024}, MAS~\cite{NascimentoAcsos2023,KannanIros2024,JiangWc2024,NiEml2024}, autotelic agents~\cite{ColasPmlr2023,ColasJair2022}, financial analysis~\cite{Finrobot2024}, and wireless networks~\cite{Wirelessagent2025}.

Though \ac{LLM}-based planning dominates, \ac{RL}-based frameworks remain popular.
In~\cite{TanNeuro2010}, beliefs, desires, and intentions are modelled via \ac{SONN}, with plans generated using \ac{RL} over the \ac{SONN} embeddings.
BDI-FALCON~\cite{TanEswa2011} and TDI-FALCON~\cite{TanTnn2008} follow similar approaches.
Meanwhile, \citeap{RamirezMates2017} rely on intentional learning \textendash{} implemented in \jason{} \textendash{} to enable \ac{RL}-based plan acquisition.
\citeap{BoselloEmas2019} apply \ac{RL} to adapt plans during the agent lifecycle, with~\cite{PulawskiAcsos2021} extending this to adversarial settings.

Other approaches use simpler \ac{ML} models for plan generation.
For example, iFalcon~\cite{SubagdjaIat2008} relies only on \ac{SONN} for plan generation and selection.
On the other hand, Neural \ac{ToM} models~\cite{RabinowitzIcml2018,KumarHci2023} use \acp{NN} to generate and select which actions to take.
\citeap{bdinl-aamas2024} propose a natural language-based \ac{BDI} agent where plans are generated from textual beliefs, desires, and intentions.
Instead, SEMLINCS~\cite{SchrodtTcs2017} uses free energy-based inference for runtime plan definition and \citeap{HoTcs2022} apply causal learning~\cite{KaddourArxiv2022} to build the agent's \ac{ToM} and generate plans.

\subsubsection{Plan selection}
Approaches in this category are more historically rooted than plan generation works.
While \ac{ML}-based plan generation gained popularity with generative \ac{AI}, plan selection typically requires simpler \ac{ML} models.
When an agent has access to a plan library, selection becomes a standard \ac{ML} classification task.
Early works like \citeap{bdilearning-ijats1} used decision trees to optimise plan selection and avoid repeated action-level failures.
Similarly,~\cite{SinghAtal2010,SinghIjcai2011} applied \ac{DT} to assess plan success likelihood, updating the tree online based on experience.

Recent approaches use more advanced models, including \acp{NN} and \ac{RL}.
For example, \citeap{WanAcai2018} apply Q-learning~\cite{WatkinsMl1992} to select plans based on estimated rewards and \citeap{SacharnyIas2021} use a similar method for unmanned aircraft systems.
In~\cite{SubagdjaIat2008,RabinowitzIcml2018,KumarHci2023,bdinl-aamas2024}, \acp{NN} map agent intentions to the most suitable plan post-generation.
Meanwhile, in~\cite{SubagdjaIat2008,RabinowitzIcml2018,KumarHci2023,bdinl-aamas2024} \acp{NN} are used to map agent intentions to the corresponding best plan after the generation process is complete.
Finally, some works rely on \ac{LLM} to either select the best plan among viable plan-like functions~\cite{Personalwab2025,Codriving2025} or generate multiple plans and concurrently select the best one, such as~\cite{chainofthoughts-neurips2022,HuangIcml2022,GramopadhyeIros2023,HaoEmnlp2023,react-iclr2023,retroformer-2024}.

\subsubsection{Plan optimisation}

Optimisation and refinement of plans at the agent level remain challenging tasks for \ac{ML} models, even with generative \ac{AI}, as plan optimisation involves identifying why a plan failed and determining the necessary patches to complete the task.
As a result, few approaches address this directly, with most focusing on replanning.
Preliminary works rely on simpler \ac{ML} models such as decision trees or genetic algorithms.
For example, \citeap{GuerraClima2004} integrate an induction paradigm into \ac{BDI} agents to optimise failed plans using logical decision trees.
Later, the same authors propose a learning-based optimisation strategy in~\cite{GuerraTas2008}, where failure data is used to generate new plans.
\citeap{ShawSaicsit2015} explore genetic algorithms for replanning in \ac{BDI} agents, showing promising results.
Finally, some recent approaches adopt advanced \ac{ML} techniques \textendash{} including \acp{LLM} and \ac{RL} \textendash{} to integrate feedback loops into the plan generation pipeline, thus enabling dynamic plan optimisation based on agent failures~\cite{chainofthoughts-neurips2022,SongIccv2023,TanNeuro2010,JiangWc2024,NiEml2024}.

\begin{mybox}
\phantomsection
{{\bf \faSearch\ \textit{Quick Takeaway:}\label{take:plan}} Planning is the most explored ML-BDI module, dominated by LLM-based plan generation; however, symbolic plan verification and integrated optimisation pipelines are still underdeveloped.}
\end{mybox}


\newcommand{\plantabhead}{
	\textbf{\#} & \textbf{Method} & \textbf{Year} & \textbf{Aim} & \textbf{ML} (\ref{item:rq4}) & \textbf{On. Learn.} (\ref{item:rq5}) & \textbf{Goal} (\ref{item:rq3}) & \textbf{Tech.} (\ref{item:rq6}) \\
	\hline\hline
}

\newcounter{PlanMethod}
\setcounter{PlanMethod}{1}
\newcommand{\newPlanMethodIndex}{\thePlanMethod\stepcounter{PlanMethod}}

\begin{scriptsize}

\begin{longtable}{c|p{3cm}|c|c|c|c|c|c}
    \caption{
        Summary of \ac{ML}-enabled planning modules in \mlbdi{} agents. Legend: O=\emph{optimization}, S=\emph{selection}, G=\emph{generation}, ILP=\emph{inductive logic programming}, DT=\emph{decision tree-like}, U=\emph{unspecified}, C=\emph{custom}, GA=\emph{genetic algorithm}, Jx=\emph{jadex}, J=\emph{jason}, E=\emph{energy-based}, A=\emph{agent-speak}, CL=\emph{causal learning}, L=\emph{language-based}.
    }
    \labeltab{plan-taxonomy}\\        
    \plantabhead
    \endfirsthead
    \caption[]{Summary of \ac{ML}-enabled planning modules in \mlbdi{} agents (continued).}\\
    \plantabhead
    \endhead
    \endlastfoot
    \newPlanMethodIndex & \citesp{GuerraClima2004} & O & ILP & \xmark & DT & \unavailable
    \\\hline
    \newPlanMethodIndex & \citesp{GuerraTas2008} & O & U & \xmark & U & \unavailable
    \\\hdashline
    \newPlanMethodIndex & \citesp{SubagdjaIat2008} & S,G & NN & \cmark & C & \unavailable
    \\\hline
    \newPlanMethodIndex & \citesp{bdilearning-ijats1} & S & DT & \xmark & U & \unavailable
    \\\hline
    \newPlanMethodIndex & \citesp{SinghAtal2010} & S & DT & \cmark & U & \unavailable
    \\\hdashline
    \newPlanMethodIndex & \citesp{TanNeuro2010} & G,O & RL & \cmark & U & \unavailable
    \\\hline
    \newPlanMethodIndex & \citesp{SinghIjcai2011} & S & DT & \cmark & U & \unavailable
    \\\hdashline
    \newPlanMethodIndex & \citesp{TanEswa2011} & G & RL & \cmark & U & \unavailable
    \\\hline
    \newPlanMethodIndex & \citesp{ShawSaicsit2015} & O & GA & \cmark & Jx & \unavailable
    \\\hline
    \newPlanMethodIndex & \citesp{RamirezMates2017} & G & RL & \xmark & J & \unavailable
    \\\hdashline
    \newPlanMethodIndex & \citesp{SchrodtTcs2017} & G & E & \xmark & C & \unavailable
    \\\hline
    \newPlanMethodIndex & \citesp{WanAcai2018} & S & RL & \xmark & A & \unavailable
    \\\hdashline
    \newPlanMethodIndex & \citesp{RabinowitzIcml2018} & S,G & NN & \xmark & C & \unavailable
    \\\hline
    \newPlanMethodIndex & \citesp{BoselloEmas2019} & G & RL & \cmark & J & \tiny{\url{https://tinyurl.com/2xr4l4bk}} 
    \\\hline
    \newPlanMethodIndex & \citesp{SacharnyIas2021} & S & RL & \xmark & U & \unavailable
    \\\hdashline
    \newPlanMethodIndex & \citesp{PulawskiAcsos2021} & G & RL & \cmark & J & \unavailable
    \\\hline
    \newPlanMethodIndex & \citesp{HoTcs2022} & G & CL & \xmark & U & \unavailable
    \\\hdashline
    \newPlanMethodIndex & \citesp{chainofthoughts-neurips2022} & G,O,S & LLM & \xmark & L & \unavailable
    \\\hdashline
    \newPlanMethodIndex & \citesp{HuangIcml2022} & G,S & LLM & \xmark & L & \tiny{\url{https://tinyurl.com/28nqxd4d}} 
    \\\hdashline
    \newPlanMethodIndex & \citesp{HuangCorl2022} & G & LLM & \xmark & L & \unavailable
    \\\hline
    \newPlanMethodIndex & \citesp{KumarHci2023} & S,G & NN & \xmark & C & \unavailable
    \\\hdashline
    \newPlanMethodIndex & \citesp{ZhuArxiv2023} & G & LLM & \xmark & L & \tiny{\url{https://tinyurl.com/26fsgvbc}} 
    \\\hdashline
    \newPlanMethodIndex & \citesp{hugginggpt-neurips2023} & G & LLM & \xmark & L & \tiny{\url{https://tinyurl.com/2dx5gvxh}} 
    \\\hdashline
    \newPlanMethodIndex & \citesp{GramopadhyeIros2023} & G,S & LLM & \xmark & L & \tiny{\url{https://tinyurl.com/28r766o9}} 
    \\\hdashline
    \newPlanMethodIndex & \citesp{HaoEmnlp2023} & G,S & LLM & \xmark & L & \tiny{\url{https://tinyurl.com/26w2s5v4}} 
    \\\hdashline
    \newPlanMethodIndex & \citesp{react-iclr2023} & G,S & LLM & \xmark & L & \tiny{\url{https://tinyurl.com/24k4begz}} 
    \\\hdashline
    \newPlanMethodIndex & \citesp{RanaCorl2023} & G & LLM & \xmark & L & \unavailable
    \\\hdashline
    \newPlanMethodIndex & \citesp{SongIccv2023} & G,O & LLM & \xmark & L & \tiny{\url{https://tinyurl.com/22coj9lw}} 
    \\\hdashline
    \newPlanMethodIndex & \citesp{NottinghamIcml2023} & G & LLM & \cmark & U & \tiny{\url{https://tinyurl.com/28555kln}} 
    \\\hdashline
    \newPlanMethodIndex & \citesp{DipaloArxiv2023} & G & LLM & \cmark & L & \unavailable
    \\\hdashline
    \newPlanMethodIndex & \citesp{WuArxiv2023} & G & LLM & \cmark & L & \unavailable
    \\\hdashline
    \newPlanMethodIndex & \citesp{XiaEtfa2023} & G & LLM & \xmark & L & \unavailable
    \\\hdashline
    \newPlanMethodIndex & \citesp{HuangArxiv2023} & G & LLM & \xmark & L & \tiny{\url{https://tinyurl.com/2d9ghoqe}} 
    \\\hdashline
    \newPlanMethodIndex & \citesp{NascimentoAcsos2023} & G & LLM & \xmark & L & \unavailable
    \\\hdashline
    \newPlanMethodIndex & \citesp{ColasPmlr2023} & G & LLM & \xmark & L & \unavailable
    \\\hdashline
    \newPlanMethodIndex & \citesp{swiftsage-neurips2023} & G & LLM & \xmark & L & \tiny{\url{https://tinyurl.com/2xkejjey}} 
    \\\hdashline
    \newPlanMethodIndex & \citesp{RuanArxiv2023} & G & LLM & \xmark & L & \unavailable
    \\\hdashline
    \newPlanMethodIndex & \citesp{GaoSthree2023} & G & LLM & \xmark & L & \unavailable
    \\\hline
    \newPlanMethodIndex & \citesp{LianWww2024} & G & LLM & \xmark & L & \tiny{\url{https://tinyurl.com/2d9ghoqe}} 
    \\\hdashline
    \newPlanMethodIndex & \citesp{voyager-tmlr2024} & G & LLM & \xmark & L & \tiny{\url{https://tinyurl.com/2dul6unn}} 
    \\\hdashline
    \newPlanMethodIndex & \citesp{bdinl-aamas2024} & S,G & NN & \xmark & L & \tiny{\url{https://tinyurl.com/27uazxbb}} 
    \\\hdashline
    \newPlanMethodIndex & \citesp{chateda-tcad43} & G & LLM & \xmark & L & \tiny{\url{https://tinyurl.com/2y8xleya}} 
    \\\hdashline
    \newPlanMethodIndex & \citesp{recmind-naacl2024} & G & LLM & \xmark & L & \unavailable
    \\\hdashline
    \newPlanMethodIndex & \citesp{embodiedagentsllm-iclr2024} & G & LLM & \xmark & L & \tiny{\url{https://tinyurl.com/29kecyhe}} 
    \\\hdashline
    \newPlanMethodIndex & \citesp{GuanKdd2024} & G & LLM & \xmark & L & \unavailable
    \\\hdashline
    \newPlanMethodIndex & \citesp{KannanIros2024} & G & LLM & \xmark & L & \tiny{\url{https://tinyurl.com/23tpb2d4}} 
    \\\hdashline
    \newPlanMethodIndex & \citesp{NiEml2024} & G,O & LLM & \xmark & L & \tiny{\url{https://tinyurl.com/29384ch2}} 
    \\\hdashline
    \newPlanMethodIndex & \citesp{JiangWc2024} & G,O & LLM & \xmark & L & \tiny{\url{https://tinyurl.com/28hclv6o}} 
    \\\hdashline
    \newPlanMethodIndex & \citesp{retroformer-2024} & G,S & LLM & \xmark & L & \tiny{\url{https://tinyurl.com/247ysg6p}} 
    \\\hdashline
    \newPlanMethodIndex & \citesp{tptuv2-2024} & G & LLM & \xmark & L & \tiny{\url{https://tinyurl.com/28ylm7gc}} 
    \\\hdashline
    \newPlanMethodIndex & \citesp{metagpt-2024} & G & LLM & \xmark & L & \tiny{\url{https://tinyurl.com/29vyobuw}} 
    \\\hdashline
    \newPlanMethodIndex & \citesp{Finrobot2024} & G & LLM & \xmark & L & \tiny{\url{https://tinyurl.com/292crmc3}} 
    \\\hline
    \newPlanMethodIndex & \citesp{ZhaoArxiv2025} & G & LLM & \cmark & L & \unavailable
    \\\hdashline
    \newPlanMethodIndex & \citesp{SchulzIjpeds2025} & G & LLM & \xmark & Jx & \unavailable
    \\\hdashline
    \newPlanMethodIndex & \citesp{CiattoEcai2025} & G & LLM & \xmark & A & \unavailable
    \\\hdashline
    \newPlanMethodIndex & \citesp{Personalwab2025} & S & LLM & \xmark & L & \tiny{\url{https://tinyurl.com/22vyf3r8}} 
    \\\hdashline
    \newPlanMethodIndex & \citesp{Wirelessagent2025} & G & LLM & \xmark & L & \tiny{\url{https://tinyurl.com/2bbt2ykd}} 
    \\\hdashline
    \newPlanMethodIndex & \citesp{Codriving2025} & S & LLM & \xmark & L & \tiny{\url{https://tinyurl.com/2a5d64mr}} 
    \\\hline\hline
    
\end{longtable}

\end{scriptsize}

\begin{figure}
\centering
\begin{minipage}{.315\textwidth}
    \centering
    \includegraphics[width=\linewidth]{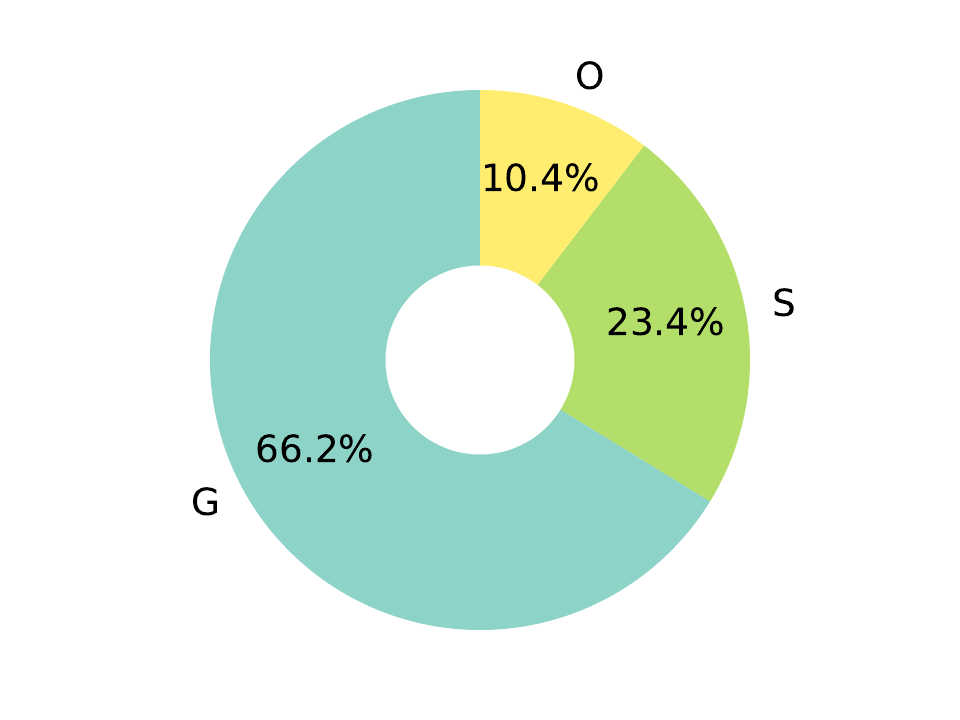}
    \captionof{figure}{Distribution of planning approaches\textemdash{}see legend of \reftab{plan-taxonomy}.}
    \labelfig{plan_aim_plot}
\end{minipage}%
\hspace{0.2cm}
\begin{minipage}{.315\textwidth}
    \centering
    \includegraphics[width=\linewidth]{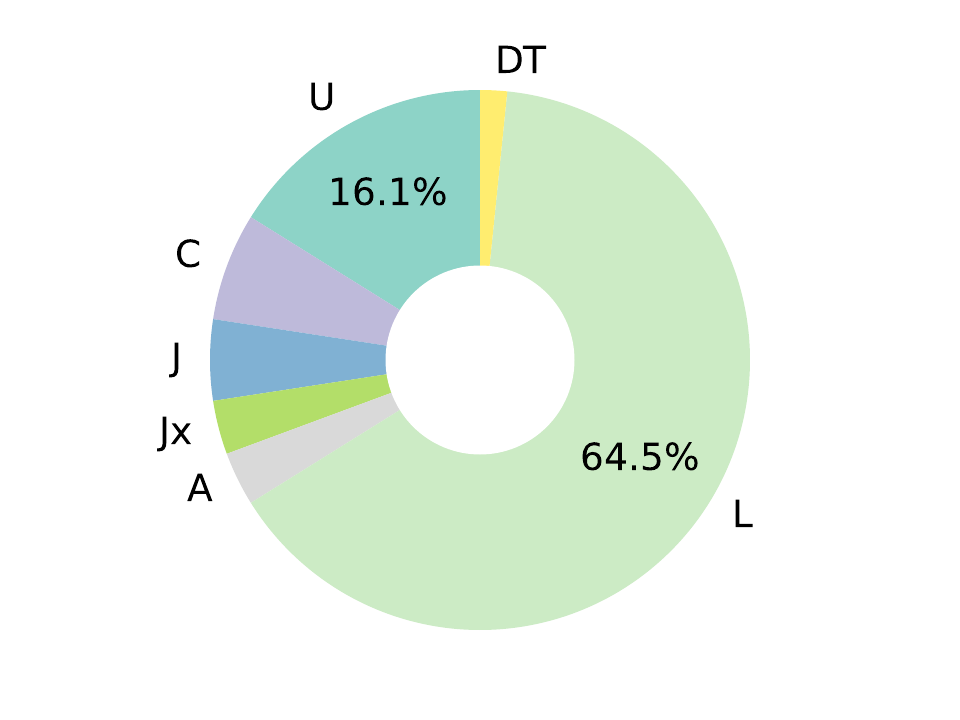}
    \captionof{figure}{Distribution of approaches to represent plans (\ref{item:rq3})\textemdash{}see legend of \reftab{plan-taxonomy}.}
    \labelfig{plan_struct_plot}
\end{minipage}
\hspace{0.2cm}
\begin{minipage}{.315\textwidth}
    \centering
    \includegraphics[width=\linewidth]{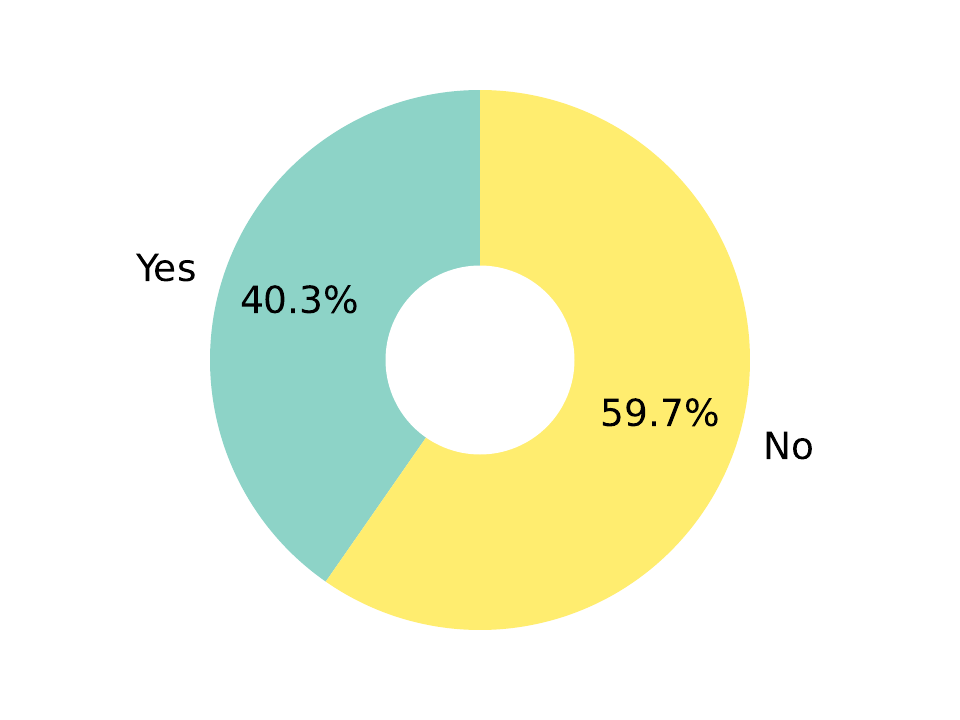}
    \captionof{figure}{Distribution of papers enabling online learning in agent planning (\ref{item:rq5}).}
    \labelfig{plan_learn_plot}
\end{minipage}
\end{figure}

\subsubsection*{\textbf{Opportunities and future directions}}

According to our analysis, \ac{ML}-based planning is the most explored integration approach in \ac{BDI} frameworks.
However, current proposals face several challenges due to their reliance on generative models and one-shot plan generation.
Most \ac{LLM}-based approaches produce textual plans that are difficult to verify without further \subsym{} processing.
In this context, generating syntactically valid plans in \sym{} or logic program form remains an open research challenge deserving more attention.
Such \sym{} plans enable formal verification~\cite{SilvaTcad2008,AvalleFac2014} and automatic detection of bugs or undesired behaviours.
Plan generation should also enforce formal syntax, similar to program synthesis~\cite{codegen-iclr2023,JainIcse2022}.

Given the widespread use of \acp{LLM} for defining agent actions, we emphasise the need for deeper analysis of their reliability.
Generative models, including \acp{LLM}, are prone to hallucinations~\cite{HuangArxiv2023Hallucinations}, which can lead to unreliable behaviour.
In \ac{BDI} agents, this translates into failed actions or violations of norms, policies, and laws.
Such risks are critical in safety-sensitive domains like healthcare~\cite{CroattiArtmed2019} and others~\cite{AdamKer2016}.
We therefore anticipate future research to focus more on the safety of \subsym{} plan generation to support real-world deployment.

Finally, as most current works propose one-shot plan generation, we highlight the lack of end-to-end frameworks that include plan generation, optimisation, and selection.
Introducing feedback loops into the planning process would help identify and correct flawed components, leading to more consistent plans.
However, designing such comprehensive planning pipelines remains a non-trivial challenge and a promising direction for future research.

\section{Machine Learning and Actions}\labelsec{ml_actions}

This section presents our findings on integrating \ac{ML} approaches into \mlbdi{} agents with respect to actions.
In a \ac{BDI} agent, actions are executed via effectors \textendash{} e.g., mechanical arms, screens \textendash{} enabling interaction with the environment.
Actions are the agent's core behaviours for achieving goals and fulfilling desires and intentions.
Our analysis identifies only a few works using \ac{ML} to support or enhance agent acting.
These typically use \acp{NN} to analyse the environment and agent status, identifying when goals are achieved through specific actions or sequences.
Such works define \subsym{} procedures for detecting successful goal completion, referred to as the \emph{goal checking} problem.
For each paper addressing this, we analyse its workflow \textendash{} addressing~\ref{item:rq1} \textendash{} and identify:
\begin{inlinelist}
    \item the \ac{ML} model used (\ref{item:rq4});
    \item the goal structure \textendash{} i.e., for which kind of goal is the agent used \textendash{} answering~\ref{item:rq3};
    \item if the proposed approach supports online learning of the \ac{ML} model inside the agent framework (\ref{item:rq5}); and
    \item if the proposed approach comes with a publicly available implementation, answering~\ref{item:rq6}.
\end{inlinelist} 

In~\cite{AmadoIjcnn2018}, autoencoders~\cite{BerahmandAir2024} are used to detect goal achievement in latent space, improving efficiency.
\citeap{PereiraIjcai2019} apply deep \acp{NN} and causal reasoning to approximate dynamic constraints in finite-horizon control, modelling agent behaviours.
Some works use large pre-trained models, including \acp{LLM} and \acp{LVM}, to track task completion.
In PET~\cite{WuArxiv2023}, an \ac{LLM}-based planner uses a question-answering module to monitor sub-task completion.
\citeap{DipaloArxiv2023} combine language and vision foundation models, using language for reasoning and vision for acting.
At each reasoning cycle, the \ac{LVM} checks sub-goal achievement and triggers plan refinement if needed.
Finally, \ac{ML} can be used not only to track the progress and predict the completion of the agent's own goals. 
For example, \citeap{ZhiNips2020} use \acp{BN} to recognise other agents' tasks and completion levels, aiming for time-effective task scheduling.

\reftab{action-taxonomy} summarises the analysed papers and their features.

\begin{mybox}
\phantomsection
{{\bf \faSearch\ \textit{Quick Takeaway:}\label{take:act}} ML is occasionally used for goal checking and action monitoring, usually coupled with ML-based planning.}
\end{mybox}


\newcommand{\actiontabhead}{
	\textbf{\#} & \textbf{Method} & \textbf{Year} & \textbf{ML} (\ref{item:rq4}) & \textbf{Online Learn} (\ref{item:rq5}) & \textbf{Goal} & \textbf{Tech.} (\ref{item:rq6}) \\
	\hline\hline
}

\newcounter{ActionMethod}
\setcounter{ActionMethod}{1}
\newcommand{\newActionMethodIndex}{\theActionMethod\stepcounter{ActionMethod}}

\begin{scriptsize}

\begin{longtable}{c|p{4cm}|c|c|c|c|c}
    \caption{
        Summary of \ac{ML}-enabled action in \mlbdi{} agents. Legend: P=\emph{puzzle completion}, N=\emph{navigation}, M=\emph{multi-type}, O=\emph{object manipulation}.
    }
    \labeltab{action-taxonomy}\\        
    \actiontabhead
    \endfirsthead
    \caption[]{Summary of \ac{ML}-enabled action in \mlbdi{} agents (continued).}\\
    \actiontabhead
    \endhead
    \endlastfoot
    \newActionMethodIndex & \citesp{AmadoIjcnn2018} & NN & \xmark & P & \unavailable
    \\\hline
    \newActionMethodIndex & \citesp{PereiraIjcai2019} & NN & \cmark & N & \unavailable
    \\\hline
    \newActionMethodIndex & \citesp{ZhiNips2020} & BN & \cmark & M & \url{https://tinyurl.com/24hzfbyl}
    \\\hline
    \newActionMethodIndex & \citesp{DipaloArxiv2023} & LVM & \xmark & O & \unavailable
    \\\hdashline
    \newDesireMethodIndex & \citesp{WuArxiv2023} & LLM & \cmark & M & \unavailable
    \\\hline\hline

\end{longtable}

\end{scriptsize}

\subsubsection*{\textbf{Opportunities and future directions}}
Given the limited number of approaches targeting \ac{ML}-enabled acting in \ac{BDI} agents and the growing focus of the \ac{AI} community on \ac{ML}, we expect increased attention to this area in future research.
Vision models can be extensively used for goal checking and environmental sensing during agent actions, supporting the development of smart effectors.
To enable complex \ac{ML}-based acting, it is essential for the \ac{BDI} research community to extend current frameworks \textendash{} such as \jason{}~\cite{jasonbook2007} and \jacamo{}~\cite{jacamo-scp78} \textendash{} to integrate learning and vision technologies like PyTorch~\cite{PaszkeNips2019} and Tensorflow~\cite{tensorflow-usenix12}.
We also envision the use of \ac{ML} for \emph{preemptive} goal reachability checks, allowing agents to halt actions when goals become unattainable due to environmental changes, thus avoiding deadlocks and resource waste.
In this context, \ac{ML} models can be used by \mlbdi{} agents to predict future environmental evolution and detect when goals will become unreachable.
We refer to this as the \emph{deadline checking} problem and consider it a promising direction for future research.

\section{Where and When to Learn?}\labelsec{learning}

In this section, we address~\ref{item:rq5} by focusing on the learning procedure in \mlbdi{} agent frameworks.
Specifically, we examine whether the proposed approaches support online learning~\cite{HoiNeuro2021} in the \mlbdi{} agent.
Online learning is essential for updating the \ac{ML} model(s) integrated into the agent.
As the environment evolves, the agent must update not only its beliefs, desires, and intentions, but also how the \ac{ML} models perform inference.
Relying on static \ac{ML} models \textendash{} trained once before deployment \textendash{} leads to issues such as concept and data drift~\cite{LuTkde2018}.
Thus, reliable \mlbdi{} agents require frameworks that support training data collection during the agent lifecycle and continuous updates of the inner \ac{ML} model(s).

As shown in \reftabfromto{sense-taxonomy}{action-taxonomy}, our survey reveals a concerning trend.
Most \mlbdi{} works lack online learning features and do not update the \ac{ML} models within the agent framework.
Of the \allcount{} papers surveyed, only \onlearningcount{} support dynamic model updates.
Several use \ac{RL} for straightforward updates~\cite{PulawskiAcsos2021,CuzzolinPsy2020}, while only a few explore fine-tuning complex models such as \ac{LLM} during the agent lifecycle~\cite{NottinghamIcml2023,DipaloArxiv2023,retroformer-2024}.
This is likely due to the complexity of iterative re-training in cognitive agents and the high computational demands of fine-tuning complex \ac{NN} models~\cite{nnconstrained-applsci11}.
Consequently, \offlearningcount{} papers adopt a static approach, training the \ac{ML} model only once before deployment.
While this may suffice for small-scale case studies with limited evaluation, it is inadequate for real-world scenarios.
Notably, \offlearnfoundcount{} out of \offlearningcount{} papers avoid online learning by relying on foundation models~\cite{ZhouArxiv2023}.
These are assumed to adapt to unseen contexts without fine-tuning.
However, this assumption fails in practice.
Although foundation models offer improved generalisation, they are not universally adaptable and still require scenario-specific updates.
We therefore stress the need for greater focus on online learning when integrating \ac{ML} techniques into \ac{BDI}(-like) agents.

\section{Conclusion}\labelsec{conclusion}

This paper surveys the state of the art of \acl{ML} integration into rational agents (\mlbdi{}), using the \acl{BDI} architecture as a reference.
We explore 20 years of literature, offering a fine-grained systematisation based on each reasoning module of a \ac{BDI} agent.
Our aim is to identify where \ac{ML} models intervene in the reasoning cycle and how they are employed.
We analyse \allcount{} primary works under the broad \mlbdi{} definition, assessing their strengths and limitations.
Findings show \ac{ML} integration into \ac{BDI} agents is a growing \ac{AI} topic, with nearly 70\% of papers published post-2020.
However, the literature is skewed: most works target belief representation and planning, often using large models \textendash{} such as \acp{LLM} \textendash{} while overlooking the critical issue of agent learning.
We hope this survey guides future research, stressing the need to prioritise agent abstractions over merely embedding \ac{ML} into increasingly complex architectures.

\section*{Acknowledgments}
This paper was partially supported by
\begin{inlinelist}
    \item ``ENGINES \textendash{} ENGineering INtElligent Systems around intelligent agent technologies'' project, funded by the European Union \textendash{} NextGenerationEU within the framework of the National Recovery and Resilience Plan NRRP \textendash{} Mission 4 ``Education and Research'' \textendash{} Component 2 \textendash{} Investment 1.1 ``National Research Program and Projects of Significant National Interest Fund (PRIN)'' \textendash{} Call PRIN 2022 \textendash{} D.D. n. 104 of 02/02/2022, under grant number 20229ZXBZM,
    \item PNRR – M4C2 – Investimento 1.3, Partenariato Esteso PE00000013 – ``FAIR—Future Artificial Intelligence Research'' – Spoke 8 ``Pervasive AI'', funded by the European Commission under the NextGenerationEU programme.
\end{inlinelist}

\bibliographystyle{abbrvnat}
\bibliography{mlbdi-short}
\end{document}